\def\year{2020}\relax
\documentclass[letterpaper]{article} 
\usepackage{aaai20}  
\usepackage{times}  
\usepackage{helvet} 
\usepackage{courier}  
\usepackage[hyphens]{url}  
\usepackage{graphicx} 
\usepackage{graphicx}  
\title{Collaborative Sampling in Generative Adversarial Networks}

\makeatletter
\newcommand{\printfnsymbol}[1]{%
  \textsuperscript{\@fnsymbol{#1}}%
}
\makeatother

\author{Yuejiang Liu\thanks{Equal contribution}, Parth Kothari\printfnsymbol{1}, Alexandre Alahi\\ 
\'Ecole Polytechnique F\'ed\'erale de Lausanne (EPFL), Switzerland\\ 
}

\usepackage{subcaption}
\usepackage[font=small,labelfont=bf]{caption}
\usepackage{amsmath}
\usepackage{amssymb}
\usepackage{algorithm}
\usepackage{algorithmic}
\usepackage{amsthm}
\usepackage{mathtools}
\usepackage{subcaption}
\usepackage{hhline}
\usepackage{array}
\usepackage{ctable} 

\newcolumntype{x}[1]{>{\centering\arraybackslash\hspace{0pt}}p{#1}}

\newcommand\szsyn{0.2}

\newcommand\szablation{0.14}

\newcommand\szcelebbox{1.74}
\newcommand\szcyclebox{0.464}

\DeclareMathOperator\supp{supp}

\begin{document}

\maketitle

\begin{abstract}
The standard practice in Generative Adversarial Networks (GANs) discards the discriminator during sampling. However, this sampling method loses valuable information learned by the discriminator regarding the data distribution. In this work, we propose a collaborative sampling scheme between the generator and the discriminator for improved data generation. Guided by the discriminator, our approach refines the generated samples through gradient-based updates at a particular layer of the generator, shifting the generator distribution closer to the real data distribution. Additionally, we present a practical discriminator shaping method that can smoothen the loss landscape provided by the discriminator for effective sample refinement. Through extensive experiments on synthetic and image datasets, we demonstrate that our proposed method can improve generated samples both quantitatively and qualitatively, offering a new degree of freedom in GAN sampling.
\end{abstract}

\section{Introduction}

Generative Adversarial Networks (GANs) \cite{Goodfellow2014GenerativeAN} are a powerful class of deep generative models known for producing realistic samples. Despite successful applications in a wide variety of tasks \cite{zhu2017unpaired,brock2018large,karras2019style}, training GANs is notoriously unstable, often impacting the model distribution. Numerous works have attempted to improve GAN training through loss functions \cite{arjovsky_wasserstein_2017}, regularization methods \cite{miyato_spectral_2018}, training procedures \cite{karras_progressive_2017} as well as model architectures \cite{radford_unsupervised_2015,zhang2019self}. Yet, stabilizing GANs at scale remains an open problem. In this work, we go beyond GAN training and explore methods for effective sampling. Our goal is to improve the model distribution by fully exploiting the value contained in the trained networks during sampling.

\begin{figure}[t]
    \centering
    \includegraphics[scale=1.0]{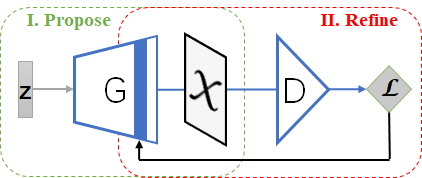}
    \caption{Once training completes, we use both the generator and the discriminator for collaborative sampling. Our scheme consists of one sample proposal step and multiple sample refinement steps. (I) The fixed generator proposes samples. (II) Subsequently, the discriminator provides gradients, with respect to the activation maps of the proposed samples, back to a particular layer of the generator. Gradient-based updates are performed iteratively.}
    \label{fig:pull_update}
\end{figure}

A standard practice in GAN sampling is to completely discard the discriminator while using only the generator for sample generation. Recent works propose to post-process the model distribution $p_g$, implicitly defined by the trained generator, using Monte Carlo techniques such as rejection sampling \cite{azadi2018discriminator} and Metropolis-Hastings independence sampler \cite{turner2019metropolis}. By rejecting undesired samples based on the output of an optimal discriminator, the accept-reject paradigm is able to recover the real data distribution $p_r$ under certain assumptions. However, these methods have several limitations:
\begin{itemize}
    \item \textit{exactness}: the assumption that the support of $p_g$ includes the support of $p_r$ is often too strong to hold in practice,
    \item \textit{efficiency}: the accept-reject procedure suffers from low sample efficiency when $p_g$ is statistically distant from $p_r$,
    \item \textit{applicability}: rejection cannot be applied to many scenarios where only one sample is produced, \textit{e.g.}, CycleGAN \cite{zhu2017unpaired}.
\end{itemize}
Drawing inspiration from Langevin \cite{roberts_exponential_1996} and Hamiltonian Monte Carlo methods \cite{Neal:1996:BLN:525544}, we address these issues by refining, rather than simply rejecting, the generated samples. 

Figure~\ref{fig:pull_update} illustrates our proposed collaborative sampling scheme between the generator and the discriminator. Once training completes, we freeze the parameters of the generator and refine the proposed samples using the gradients provided by the discriminator. This gradient-based sample refinement can be performed repeatedly at any layer of the generator, ranging from low-level feature maps to the final output space, until the samples look ``realistic" to the discriminator.

The performance of our collaborative sampling scheme is dependent on the loss landscape provided by the discriminator. To further improve the sample refinement process, we propose a practical discriminator shaping method that fine-tunes the discriminator using the refined samples. This shaping method not only enhances the robustness of the discriminator for classification but also smoothens the learned loss landscape, thereby strengthening the discriminator's ability to guide the sample refinement process. 

Our sample refinement method is not mutually exclusive with the accept-reject paradigm. An additional rejection step can be applied subsequent to the refinement process for distribution recovery. To ensure the effectiveness of the rejection step, we propose to diagnose the optimality of the discriminator with the Brier Score \cite{brier1950verification} in contrast to the calibration measure used in \cite{turner2019metropolis}. 

Through experiments on a synthetic imbalanced dataset where the standard GAN training is prone to mode collapse, we first show that the previous accept-reject methods may fail due to their strict assumptions, whereas our proposed method achieves superior results on both quality and diversity. We further demonstrate that our method can scale to the image domain effectively and provide consistent performance boost across different models including DCGAN \cite{radford_unsupervised_2015}, CycleGAN \cite{zhu2017unpaired} and SAGAN \cite{zhang2019self}. Our proposed method can be applied on top of existing GAN training techniques, offering a new degree of freedom to improve the generated samples. Code is available online\footnote{https://github.com/vita-epfl/collaborative-gan-sampling}. 


\section{Background}

\subsection{Generative Adversarial Networks}

Generative Adversarial Networks (GANs) \cite{Goodfellow2014GenerativeAN} consist of two neural networks, namely the generator G and the discriminator D, trained together. The role of the generator $G$ is to transform a latent vector ${z}$ sampled from a given distribution ${p_z}$ to a realistic sample ${G(z)}$, whereas the discriminator $D$ aims to tell whether a sample comes from the generator distribution ${p_g}$ or the real data distribution ${p_r}$. Training GANs is essentially a minimax game between these two players: 
\begin{equation*} 
    \min_G \max_D \mathbb{E}_{x \sim p_r} [ \log(D(x)) ] + \mathbb{E}_{z \sim p_z} [1 - \log(D(G(z))].
\end{equation*} 
The solution of this optimization problem, under certain conditions, leads to a generator capable of modelling the data distribution ${p_r}$. However, training GANs is notoriously unstable in practice due to the complex dynamics of the minimax game. Our goal is to sidestep the training issues and improve the generated samples during the sampling process. 

\subsection{Monte Carlo Methods}

Monte Carlo (MC) methods \cite{fermi1948note} are a broad family of algorithms which aim to draw a set of i.i.d. samples from a target distribution $p(x)$. When it is hard to directly sample from $p(x)$, one class of MC algorithms are invented to first sample from another easy-to-sample proposal distribution $q(x)$ and subsequently reject some through an accept-reject scheme. Rejection sampling and Metropolis-Hastings independence sampling \cite{tierney1994markov} are two such instances. Rejection sampling draws random samples from $q(x)$ and accepts them with probability $A(x) = p(x) / M q(x)$ if there exists an $M < \infty$ such that $p(x) \le M q(x)$ for all $x$. The Metropolis-Hastings independence sampler compares a new sample $y$ with the current one $x$, accepting $y$ with probability $A(x,y) = \text{min} \{1, {p(x)q(y)} / {p(y)q(x)} \}$. If the support of $q(x)$ includes the support of $p(x)$, these accept-reject methods are guaranteed to converge to the target distribution. However, their efficiency highly depends on the statistical distance between $q(x)$ and $p(x)$.

For the cases where it is practically difficult to find a good proposal $q(x)$, more sophisticated MC methods that leverage informed local moves to explore the important regions of $p(x)$ are preferred for efficiency. Popular algorithms include Langevin \cite{roberts_exponential_1996} and Hamiltonian MC \cite{Neal:1996:BLN:525544}, which incorporate the information of the target distribution, in the form of $\nabla \log p(x)$, to construct a gradient-based Markov transition $T(x \rightarrow y)$. A sufficient condition for an \textit{ergodic} Markov chain to converge to the target distribution $p(x)$ is \textit{reversibility}, also known as the detailed balance condition, $p(x)T(x \rightarrow y) = p(y)T(y \rightarrow x)$. 

\subsection{GAN Sampling}

The standard GAN sampling process \cite{Goodfellow2014GenerativeAN} draws samples from the generator without the involvement of the discriminator. Recently, \cite{azadi2018discriminator} proposed a rejection sampling scheme which uses the discriminator to filter out samples that are unlikely to be real. In the ideal setting, computing the acceptance probability is made possible by an optimal discriminator $D^*(x)$ as it yields the density ratio between the target $p_r(x)$ and proposal $p_g(x)$: 
\begin{equation}
    \frac{p_r(x)}{p_g(x)} = \frac{D^*(x)}{1-D^*(x)}
\end{equation}
Another recent work \cite{turner2019metropolis} proposed to replace the rejection sampling by Metropolis-Hastings independence sampling, leveraging the same knowledge about the density ratio to scale better in high dimensions. 

Nevertheless, both these methods rely on an accept-reject principle that inevitably sacrifices \textit{sample efficiency}, \textit{i.e.}, a significant number of generated samples are rejected, and \textit{flexibility}, \textit{i.e.}, the accepted samples are restricted to the data manifold learned by the generator. Our work explores a more involved collaboration scheme between the generator and the discriminator, which exploits the gradient of the density ratio provided by the discriminator to modify the generated samples. 


\section{Method}

In this section, we describe our collaborative sampling method in GANs that uses both the generator and the discriminator to produce samples (at test time). Subsequently, we introduce a discriminator shaping method that smoothens the loss landscape to enhance the effectiveness of our proposed scheme. 

\subsection{Collaborative Sampling}

Consider a generator network that inputs a latent code $z \in \mathbb{R}^{m}$ and produces an output $x \in \mathbb{R}^{n}$. It typically consists of multiple layers: 
\begin{equation}
\begin{split}
& G(z) = G_L \circ G_{L-1} \circ \dots \circ G_{1}(z), \\
& G_l(x_l) = \sigma (\theta_l \cdot x_l) + b_l, \quad l = 1, 2, \dots, L, 
\end{split}
\end{equation}
where $G_l$ is the $l$th layer of the generator, $x_l$ is the corresponding activation input, $\sigma$ is a nonlinear activation function, $\theta_l$ and $b_l$ are the model parameters. The input to the first layer is $x_1 = z$ and the output of the last layer is $G_L(x_L)=x$. For a randomly drawn sample from the generator distribution, \textit{i.e.}, $x \sim {p_{g}}$, the discriminator outputs a real-valued scalar $D(x)$ which indicates the probability of $x$ to be real. When the generator and the discriminator reach an equilibrium, the generated samples are no longer distinguishable from the real samples, \textit{i.e.}, $D^*(x) = \frac{p_r(x)}{p_r(x)+p_g(x)} = 1/2$. However, such a saddle point of the minimax problem is hardly obtained in practice \cite{arora2017generalization}, indicating room for improvement over the model distribution $p_g$. 

\begin{algorithm}[t]
\caption{Collaborative Sampling} \label{alg:sampling}
\begin{algorithmic}[1]
\STATE {\textbf{Input:} a frozen generator $G$, a frozen discriminator $D$, the layer index for sample refinement $l$, the maximum number of steps $K$, the stopping criterion $\eta$} 
\STATE {\textbf{Output:} a synthetic sample $x$} 
\STATE {Randomly draw a latent code $z$} 
\STATE {$x^0 \leftarrow$ ProposeSample($G, z$)} 
\FOR {$k = 0, 1, \dots, K-1$} 
   \IF {$D(x^k) < \eta$} 
   \STATE {$g_l^{k} \leftarrow$ GetGradient($D,x_l^k$)},
   \STATE {$x_l^{k+1} \leftarrow$ UpdateActivation($g_l^k,x_l^k$)}, \quad (Eq.~\ref{eq:backward}) 
   \STATE {$x^{k+1} \leftarrow$ UpdateSample($G,x_l^{k+1}$)}, \quad (Eq.~\ref{eq:forward}) 
   \ELSE 
   \STATE {break} 
   \ENDIF 
\ENDFOR
\end{algorithmic}
\end{algorithm} 

Our goal is to shift $p_g$ towards $p_r$ through sampling without changing the parameters of the generator. Inspired by the gradient-based MC methods using Langevin \cite{roberts_exponential_1996} or Hamiltonian \cite{Neal:1996:BLN:525544} dynamics, we leverage the gradient information provided by the discriminator to continuously refine the generated samples through the following iterative updates:
\begin{equation} \label{eq:backward}
x_l^{k+1} = x_l^{k} - \lambda \nabla_l \mathcal{L}_{G} (x_l^k),
\end{equation}
\begin{equation}\label{eq:forward}
x^{k+1} = G_L \circ G_{L-1} \circ \dots G_{l}(x_l^{k+1}),
\end{equation}
where $k$ is the iteration number, $\lambda$ is the stepsize, $l$ is the index of the generator layer for sample refinement, $\mathcal{L}_{G}$ is the loss of the generator, \textit{e.g.}, the non-saturating loss advocated in \cite{Goodfellow2014GenerativeAN}:
\begin{equation}
    \mathcal{L}_{G} = - \mathbb{E}_{z \sim p_{z}}[\log D(G(z))]
\end{equation}
The iterative sample update consists of two parts: in the backward pass, the discriminator provides the generator with gradient feedback to adjust the activation map of the selected layer $l$ (Eq.~\ref{eq:backward}); in the forward pass, the generator reuses part of its parameters to propose an improved sample (Eq.~\ref{eq:forward}). A pseudo code is summarized in Algorithm~\ref{alg:sampling}. 

Recall that an optimal discriminator outputs the density ratio between $p_r(x)$ and $p_g(x)$. The iterative updates shift samples to the regions in which $p_r(x)/p_g(x)$ is higher. In other words, samples are encouraged to move to regions where less samples are produced by the generator but more samples are expected in the real data distribution. Our method forms a closed-loop sampling process, allowing both the generator and the discriminator to contribute to sample generation. 

\begin{algorithm}[t]
\caption{Discriminator Shaping} \label{alg:shaping}
\begin{algorithmic}[1]
\STATE {\textbf{Input:} a frozen generator $G$, a pre-trained discriminator $D$, the batch size $m$}
\STATE {\textbf{Output:} a fine-tuned discriminator $\tilde{D}$}
\FOR {number of D shaping iterations}
   \STATE {Draw $m$ refined samples $\{x_c^{(1)}, \dots, x_c^{(m)}\}$ from the collaborative data distribution $p_c(x)$ according to Algorithm 1} 
   \STATE {Draw $m$ real samples $\{x_r^{(1)}, \dots, x_r^{(m)}\}$ from the real data distribution $p_r(x)$} 
   \STATE {Shape the discriminator by minimizing the objective function Eq.~\ref{eq:finetune}}
\ENDFOR
\end{algorithmic}
\end{algorithm}

\begin{figure*}[t!]
    \centering
    \begin{subfigure}[t]{0.38\columnwidth}
        \centering
        \includegraphics[width=0.95\columnwidth]{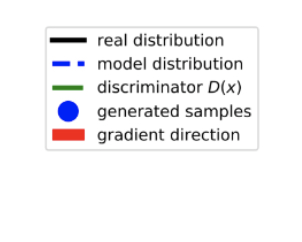}
    \end{subfigure}
    \begin{subfigure}[t]{0.38\columnwidth}
        \centering
        \includegraphics[width=0.95\columnwidth]{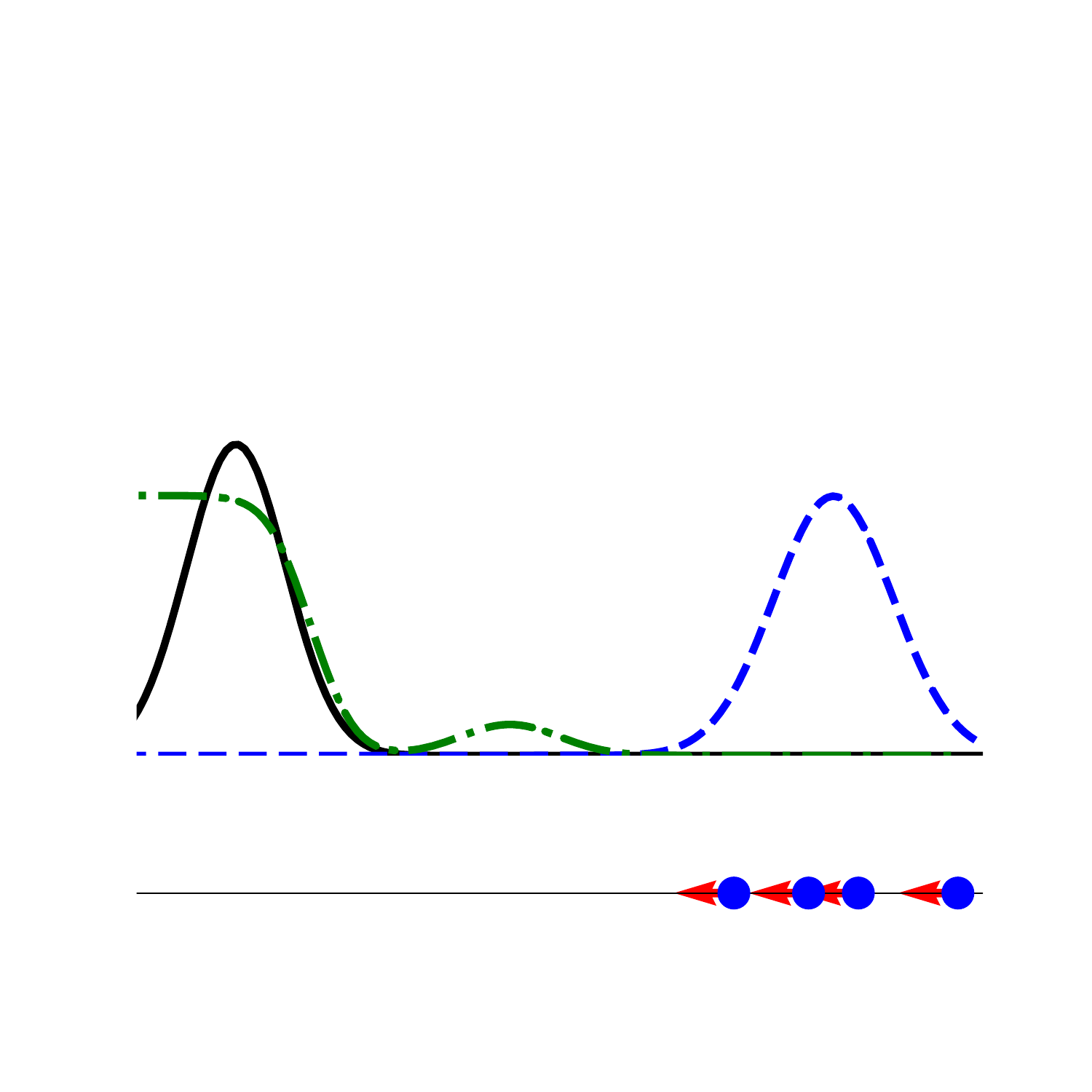}
        \caption{}
        \label{fig:far}
    \end{subfigure}
    \begin{subfigure}[t]{0.38\columnwidth}
        \centering
        \includegraphics[width=0.95\columnwidth]{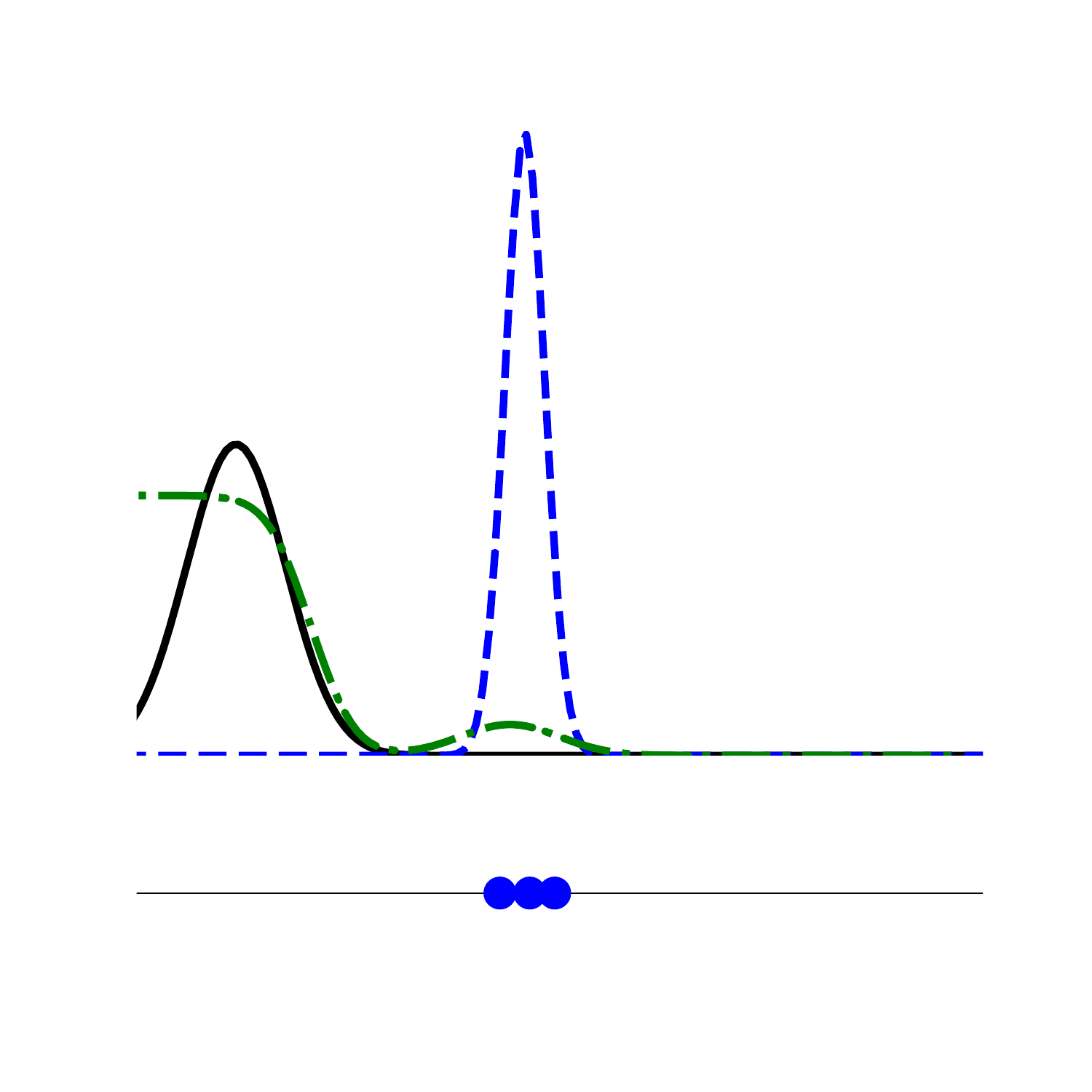}
        \caption{}
        \label{fig:stuck}
    \end{subfigure}
    \begin{subfigure}[t]{0.38\columnwidth}
        \centering
        \includegraphics[width=0.95\columnwidth]{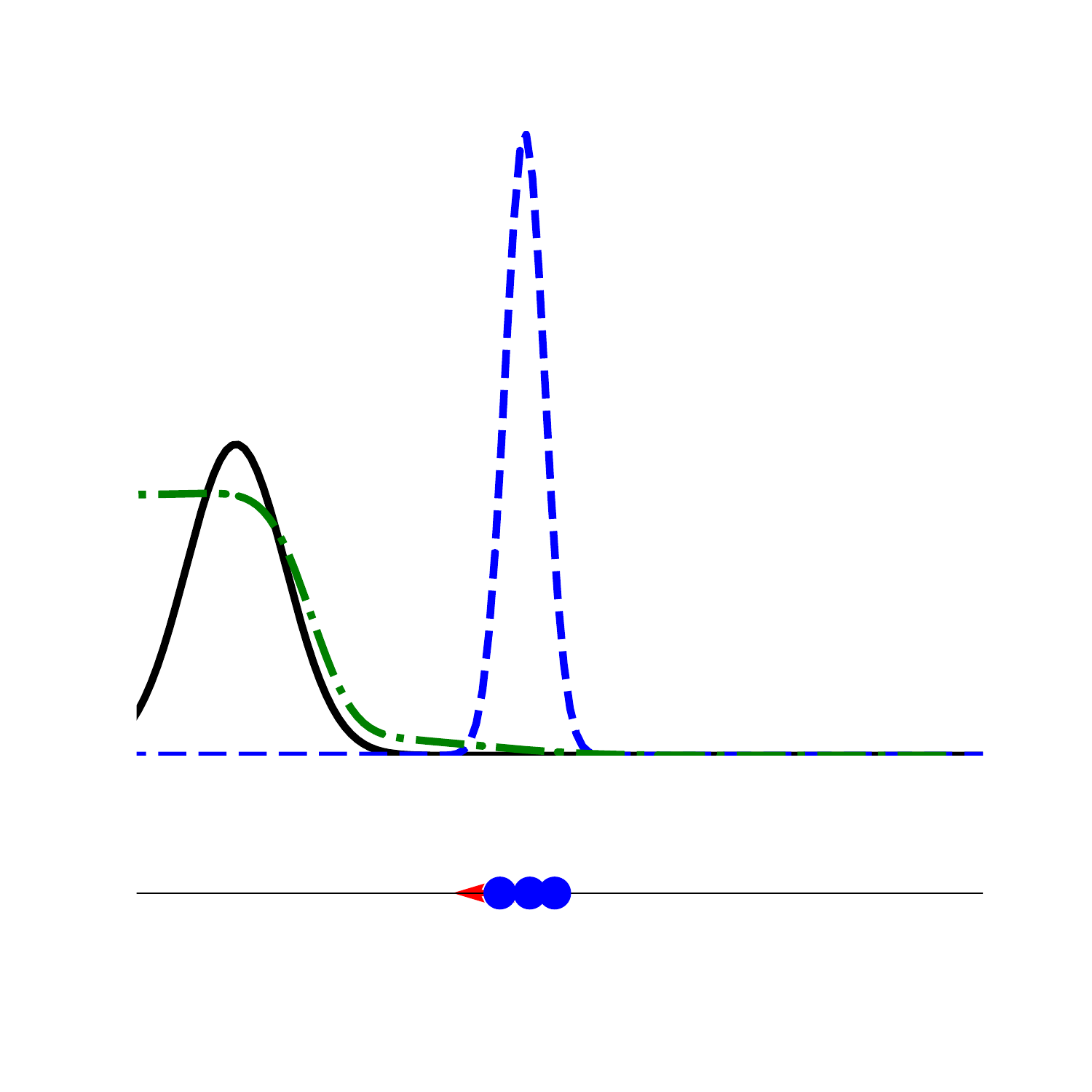}
        \caption{}
        \label{fig:shape}
    \end{subfigure}    
    \begin{subfigure}[t]{0.38\columnwidth}
        \centering
        \includegraphics[width=0.95\columnwidth]{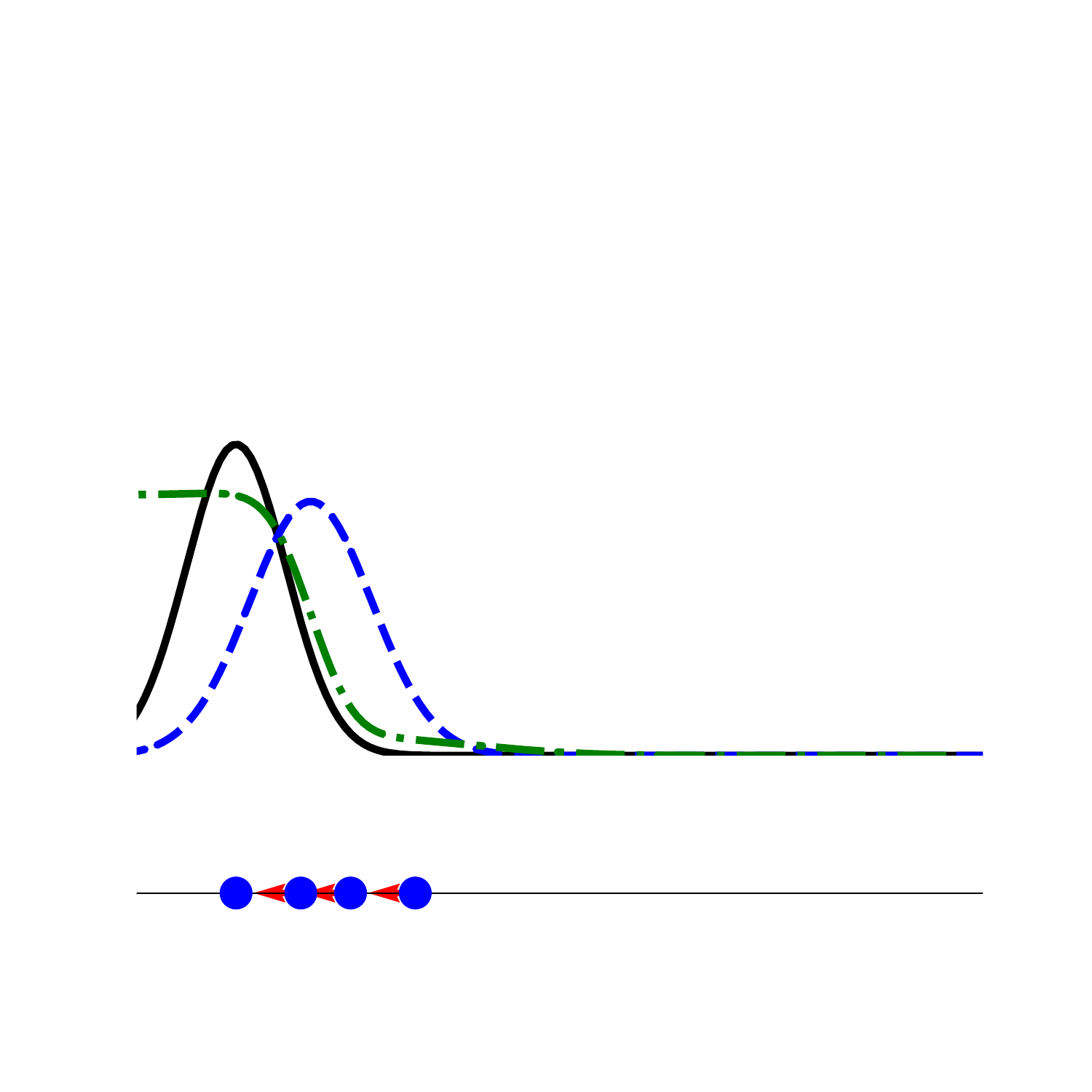}
        \caption{}
        \label{fig:final}
    \end{subfigure}
    \caption{Illustration of our collaborative sampling scheme with discriminator shaping. (a) The trained generator implicitly provides a model distribution that is close to, but not identical to, the real data distribution. At this stage, the gradient provided by the discriminator suggests informed moves. (b) However, the loss landscape from the discriminator may present local optima, hence the sample refinement process ceases. (c) Our discriminator shaping method uses the refined samples to smoothen the loss landscape. (d) The shaped loss landscape is able to better guide the refinement process, shifting the model distribution closer to the target.}
    \label{fig:illustration}
\end{figure*} 
\subsection{Discriminator Shaping} \label{sec:shaping}

In the ideal scenario where the loss landscape is smooth and monotonic from $p_g$ to $p_r$, the generated samples can be easily refined towards the real ones according to the gradient feedback. However, this is not always the case in practice for two reasons: 
\begin{itemize}
    \item In the standard GAN training, the objective of the discriminator is solely to distinguish the real and fake samples. This makes the discriminator prone to overfitting to the generator distribution and less robust in unexplored regions. 
    \item When $p_g$ fails to provide good coverage of the target $p_r$, the density ratio $p_r(x)/p_g(x)$ grows dramatically in the regions where $p_g(x) \rightarrow 0$ and $p_r(x) > 0$, giving the discriminator a false sense of $x$ being highly realistic even when $p_r(x)$ is very small. 
\end{itemize}
As a consequence, the discriminator obtained from standard training may misclassify a poorly refined sample as real and fail to suggest further improvements. 

To resolve this issue, we devise a practical discriminator shaping method, the goal of which is to strengthen the discriminator such that it is not only accurate in classifying the generated samples but also capable of effectively guiding the sample refinement process. Given the trained generator and discriminator, we fine-tune the discriminator using the refined samples: 
\begin{equation}
\mathcal{L}_{D} = - \mathbb{E}_{x \sim p_r}[\log D(x)] - \mathbb{E}_{x' \sim p_{c}}[1 - \log D(x')], \label{eq:finetune}
\end{equation}
where $x'$ is a refined sample and $p_c$ is the refined data distribution obtained from our collaborative sampling scheme. 

As outlined in Algorithm~\ref{alg:shaping}, we conduct the discriminator shaping and collaborative sampling alternatively. This post-training procedure gradually expands the coverage of the model distribution and enforces the discriminator to generalize and better collaborate with the generator for sample refinement. Figure~\ref{fig:illustration} illustrates the proposed method in a simple 1D scenario, showing how the discriminator shaping method using the refined data can help in better approximating the real data distribution. 

\subsection{Discussion}

\subsubsection{Termination Condition}

The stopping criterion $\eta$ in Algorithm~\ref{alg:sampling} can be constructed either deterministically or probabilistically depending upon the objective of the application. In cases where only sample quality matters, \textit{e.g.}, image manipulation, setting $\eta$ to the median of the discriminator outputs for real samples is a good strategy. On the other hand, when sample diversity is of significant interest, the termination condition can be defined in a probabilistic manner, \textit{e.g.}, stopping the sample refinement process at each step with positive probability. This design choice expands the support of the model distribution $\supp (p_g) \subseteq \supp(p_c)$.

\subsubsection{Rejection Step}

To recover the exact target distribution from the refined model distribution, the acceptance probability of a new refined sample needs to satisfy the detailed balance condition: 
\begin{equation}
A(\hat{x},\hat{y}) = \text{min} \Big\{1, \frac{p_r(\hat{y})}{p_r(\hat{x})} \frac{q(\hat{x}|\hat{y})}{q(\hat{y}|\hat{x})}\Big\},
\end{equation}
where $\hat{(\cdot)}$ denotes a refined sample obtained from Algorithm~\ref{alg:sampling}, $\hat{x}$ is the currently accepted sample, $\hat{y}$ is the new refined sample, $q(\hat{y}|\hat{x})$ is the transition probability. While the original samples $x$ and $y$ are independently produced by the generator, the refined ones $\hat{x}$ and $\hat{y}$ are no longer independent due to the shared loss function as well as the generator parameters between the respective refinement processes. However, when the state space is sufficiently large, the refinement trajectories $x \rightarrow \hat{x}$ and $y \rightarrow \hat{y}$ have negligible probability of overlap. Under the assumption that $\hat{x}$ and $\hat{y}$ are independent, we approximate the acceptance probability as in \cite{turner2019metropolis}:
\begin{equation}
\begin{split}
    A(\hat{x},\hat{y}) & \approx \text{min} \Big\{1, \frac{p_r(\hat{y})}{p_r(\hat{x})} \frac{q(\hat{x})}{q(\hat{y})} \Big\}, \\
    & = \text{min} \Big\{1, \frac{D^*(\hat{y})}{D^*(\hat{x})} \frac{1-D^*(\hat{x})}{1-D^*(\hat{y})} \Big\}. 
\end{split}
\end{equation}

\subsubsection{Discriminator Diagnosis}

The accept-reject procedure allows for recovering the target distribution only when the discriminator is optimal. However, it is non-trivial to obtain such a discriminator in practice. Previous work \cite{turner2019metropolis} proposes to calibrate the trained discriminator and diagnose it with the Z-statistic \cite{dawid1997prequential}
\begin{equation}
    Z = \frac{\sum_{i=1}^N y_i - D(x_i)}{\sqrt{\sum_{i=1}^N D(x_i) (1 - D(x_i) )}}
\end{equation}
where $y_i$ is the label of sample $i$, $N$ is the number of samples in the test set. 


While having reliable confidence estimates is a necessary condition for the discriminator to reach optimality, it is far from sufficient. One simple counter example is that a random binary classifier is perfectly calibrated on a testset containing an equal amount of real and generated samples, even though it performs poorly in the classification problem. To address this issue, we assess the optimality of the discriminator using the Brier Score \cite{brier1950verification}:
\begin{equation}
    BS = \frac{1}{N} \sum_{i=1}^N (y_i - D(x_i))^2
\end{equation}
The Brier Score can be decomposed into three terms including not only reliability (calibration) but also resolution and uncertainty \cite{murphy1973new}, thereby measuring the optimality of the discriminator in a broader sense. We employ the Brier Score in the diagnosis of the discriminator before performing the accept-reject step.


\begin{figure*}[t]
    \centering
    \begin{subfigure}[b]{0.29\columnwidth}
        \centering
        \includegraphics[scale=\szsyn]{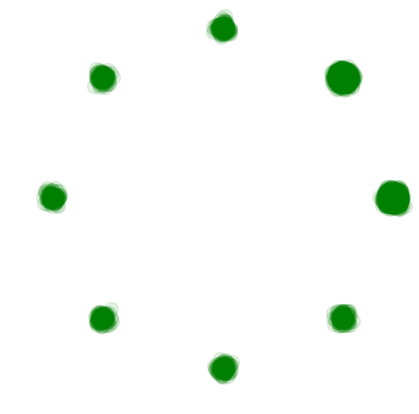}
    \end{subfigure}
    \begin{subfigure}[b]{0.29\columnwidth}
        \centering
        \includegraphics[scale=\szsyn]{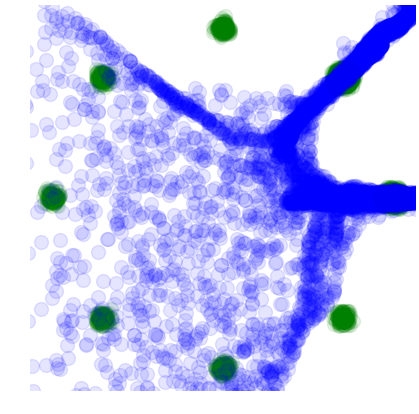}
    \end{subfigure}
    \begin{subfigure}[b]{0.29\columnwidth}
        \centering
        \includegraphics[scale=\szsyn]{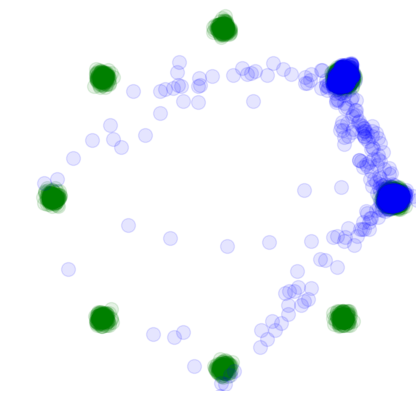}
    \end{subfigure}    
    \begin{subfigure}[b]{0.29\columnwidth}
        \centering
        \includegraphics[scale=\szsyn]{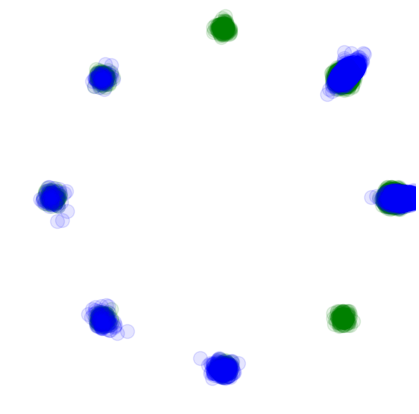}
    \end{subfigure}
    \begin{subfigure}[b]{0.29\columnwidth}
        \centering
        \includegraphics[scale=\szsyn]{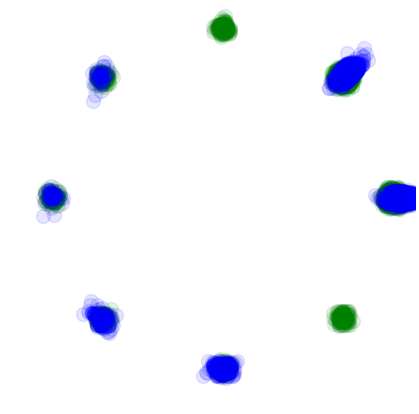}
    \end{subfigure}
    \begin{subfigure}[b]{0.29\columnwidth}
        \centering
        \includegraphics[scale=\szsyn]{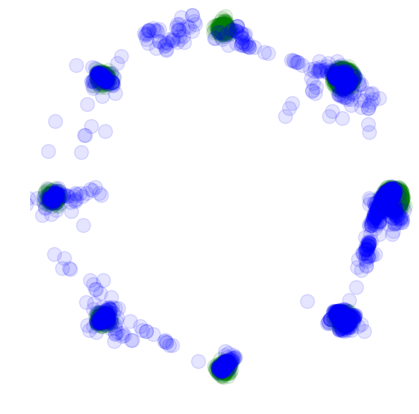}
    \end{subfigure}
    \begin{subfigure}[b]{0.29\columnwidth}
        \centering
        \includegraphics[scale=\szsyn]{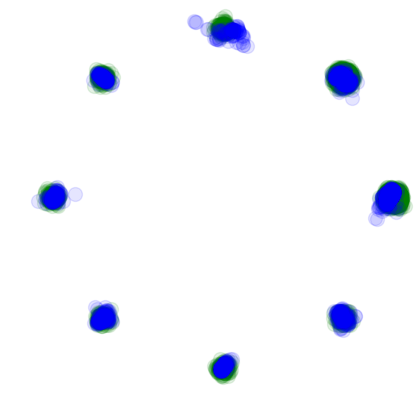}
    \end{subfigure}
    \\
    \begin{subfigure}[b]{0.29\columnwidth}
        \centering
        \includegraphics[scale=\szsyn]{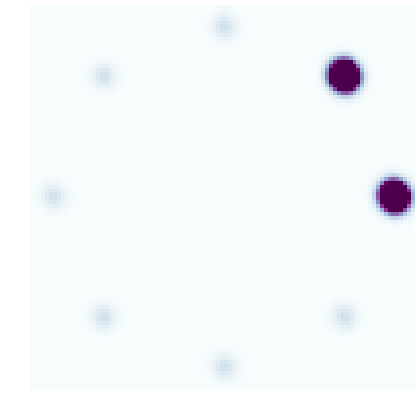}
        \caption{Real}
        \label{fig:l_reject}
    \end{subfigure}  
    \begin{subfigure}[b]{0.29\columnwidth}
        \centering
        \includegraphics[scale=\szsyn]{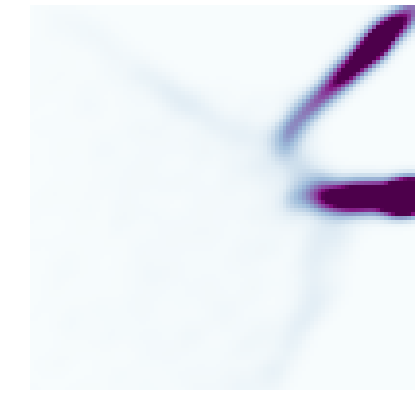}
        \caption{GAN-1k}
        \label{fig:l_mh}
    \end{subfigure}  
    \begin{subfigure}[b]{0.29\columnwidth}
        \centering
        \includegraphics[scale=\szsyn]{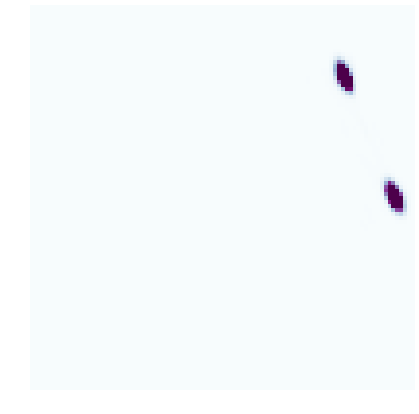}
        \caption{GAN-9k}
        \label{fig:l_origin}
    \end{subfigure}    
    \begin{subfigure}[b]{0.29\columnwidth}
        \centering
        \includegraphics[scale=\szsyn]{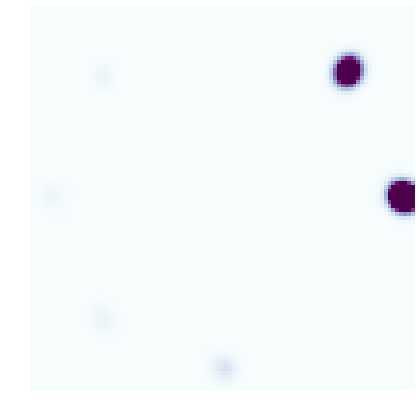}
        \caption{Reject-1k}
        \label{fig:l_cgs_1}
    \end{subfigure}
    \begin{subfigure}[b]{0.29\columnwidth}
        \centering
        \includegraphics[scale=\szsyn]{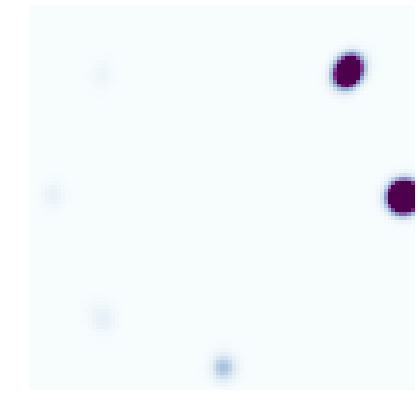}
        \caption{MH-1k}
        \label{fig:l_cgs_2}
    \end{subfigure}
    \begin{subfigure}[b]{0.29\columnwidth}
        \centering
        \includegraphics[scale=\szsyn]{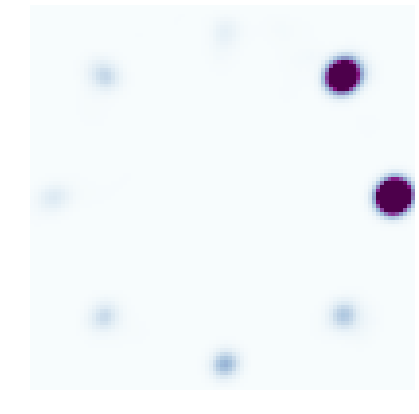}
        \caption{Refine-1k}
        \label{fig:l_cgs_3}
    \end{subfigure}     
    \begin{subfigure}[b]{0.29\columnwidth}
        \centering
        \includegraphics[scale=\szsyn]{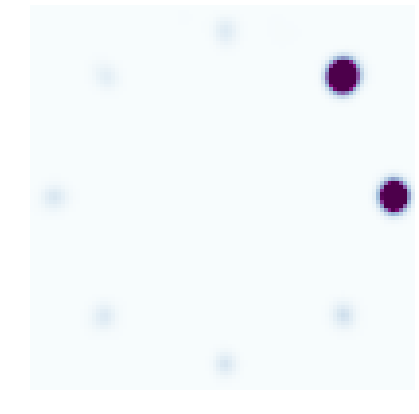}
        \caption{Collab-1k}
        \label{fig:l_cgs_4}
    \end{subfigure}         
    \caption{Qualitative evaluation of collaborative sampling in GANs on a synthetic imbalanced mixture of eight Gaussians \textit{(green)}. We draw 10k samples \textit{(first row)} from different models and visualize the resulting model distribution using kernel density estimation (KDE) \textit{(second row)}. The output samples \textit{(blue)} from the generator at an early stage of training are not of good quality \textit{(b)}, whereas training GANs longer results in mode collapse \textit{(c)}. Our sample refinement method \textit{(f)} applied to the early terminated GAN not only shifts the proposed samples closer to the real Gaussian components but also expands the categorical coverage. By incorporating the rejection step, \textit{(g)} the full version of our collaborative sampling scheme succeeds in recovering all modes without compromising sample quality, significantly outperforming \textit{(d)} the rejection sampling \cite{azadi2018discriminator} and \textit{(e)} the Metropolis-Hastings (MH) algorithm with independence sampler \cite{turner2019metropolis}.}
    \label{fig:2D}
\end{figure*}

\begin{figure*}[ht!]
    \centering
    \begin{subfigure}[b]{0.495\columnwidth}
        \centering
        \includegraphics[width=0.95\columnwidth]{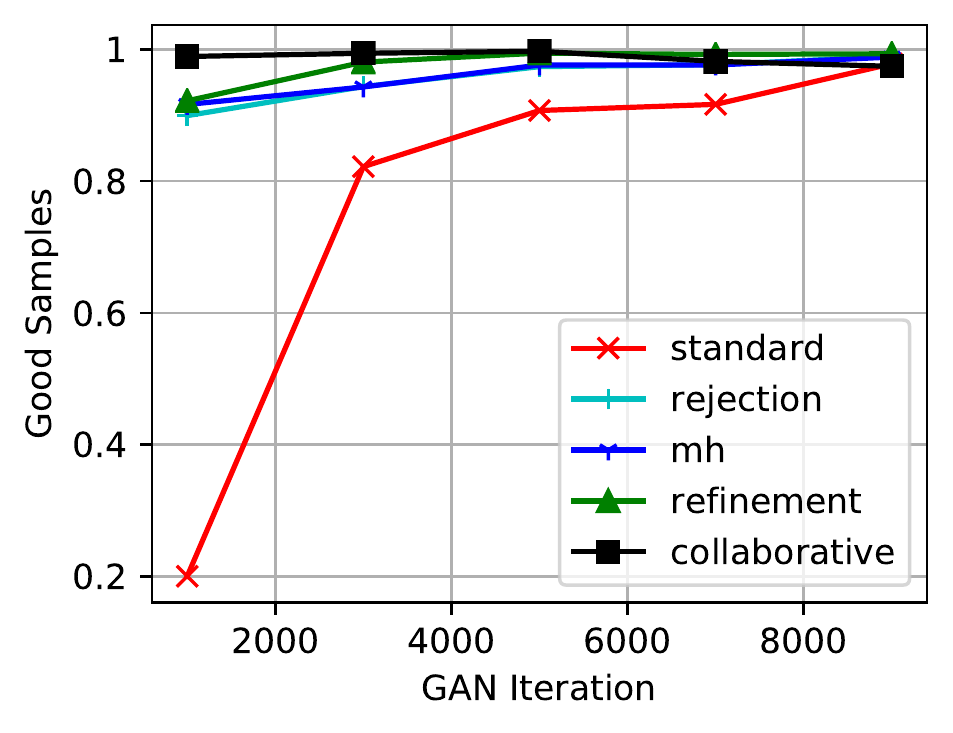}
        \caption{Quality}
    \end{subfigure}
    \begin{subfigure}[b]{0.495\columnwidth}
        \centering
        \includegraphics[width=0.95\columnwidth]{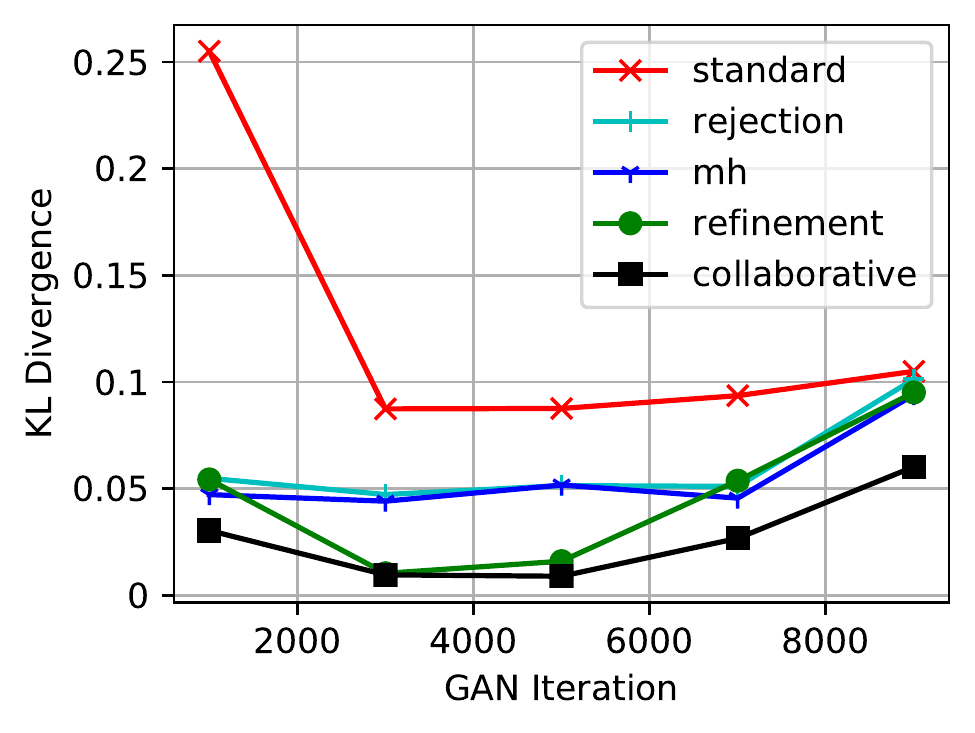}
        \caption{Diversity}
    \end{subfigure}
    \begin{subfigure}[b]{0.495\columnwidth}
        \centering
        \includegraphics[width=0.95\columnwidth]{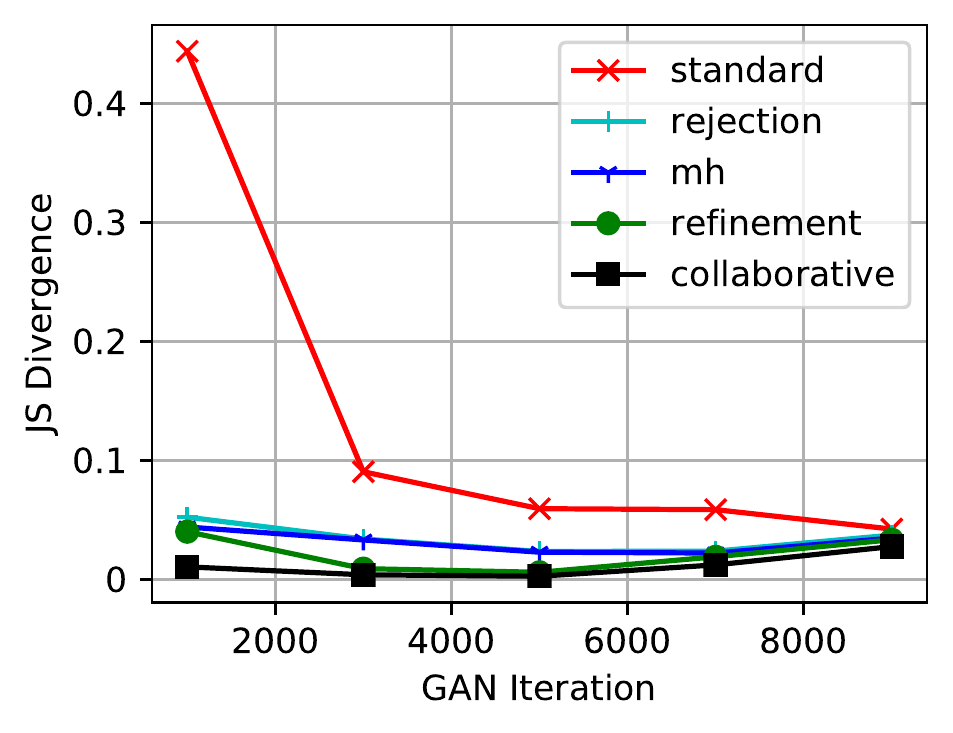}
        \caption{Overall}
    \end{subfigure}
    \begin{subfigure}[b]{0.495\columnwidth}
        \centering
        \includegraphics[width=0.95\columnwidth]{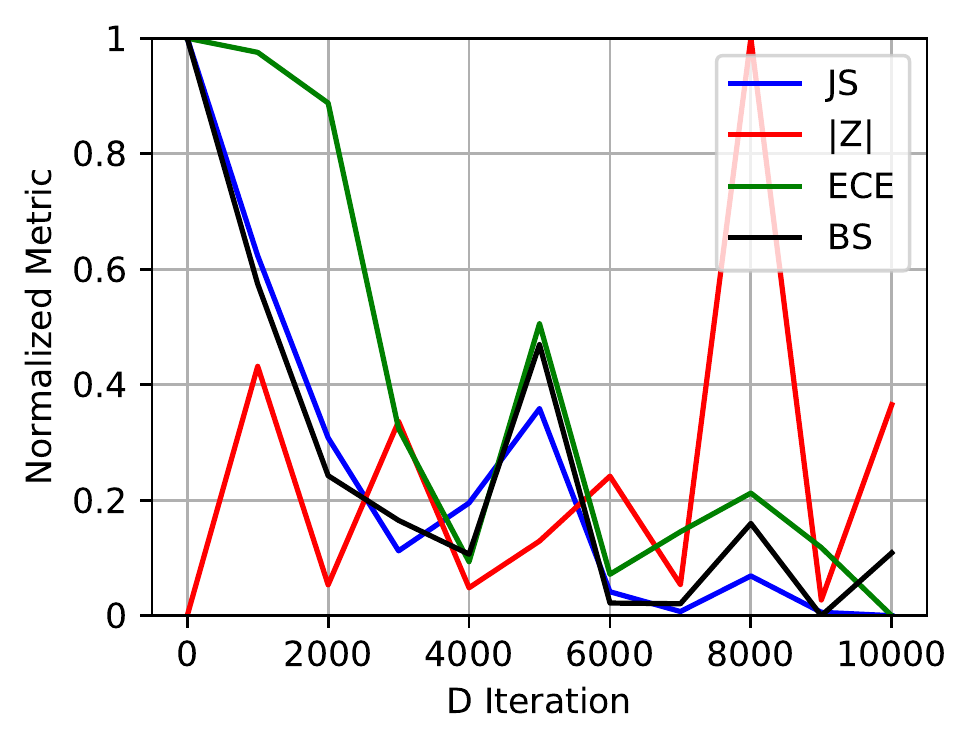}
        \caption{Diagnosis}
        \label{fig:diagnosis}
    \end{subfigure}
    \caption{Quantitative results on the imbalanced mixture of eight Gaussians. Evaluated on 10000 samples. \textit{(a)} Proportion of good samples. Higher is better. \textit{(b)} KL divergence between the categorical distribution of real samples and that of good generated samples. Lower is better. \textit{(c)} JS divergence between the augmented categorical distribution of real samples and that of all generated samples. Lower is better. \textit{(d)} The scores of diagnostic metrics and the performance gain from the rejection step at different stages of the discriminator. We apply the MH method to the generator at 1k and normalize the results on each metric to [0,1] for comparison. Among the three diagnostic metrics, the evolvement of the Brier Score exhibits the strongest similarity to that of the JS divergence.}
    \label{fig:2D_eval}
\end{figure*}

\subsubsection{Refinement Layer}

Another key hyperparameter in our method is the index of the generator layer $l$ for sample refinement. On one extreme, we can adjust the proposed sample at the output of the generator, which is equivalent to modifying the sample directly. Manipulating a proposed sample in the data space does not rely on any part of the generator and thus can, in principle, result in an optimal refinement without any constraints. However, shifting a high-dimensional sample from a low-density region to a high-density region in the data space often requires a large number of iterations. On the other extreme, we can choose to adjust the latent code $z$. As the dimension of the latent space is typically much smaller, one can obtain higher computational efficiency. However, this choice restricts the refined samples to the data prior learned by the generator and undermines the assumption of independence between $\hat{x}$ and $\hat{y}$. We empirically find that refining a sample at a middle layer of the generator leads to a good balance between efficiency and flexibility. 

\subsubsection{Computational Expenses}

Our collaborative sampling scheme provides higher sample quality at the expense of extra iterations. The additional computational cost not only depends on the choice of the refinement layer and the optimization algorithm but also reflects the quality gap between the proposed samples and refined ones. In the next section, we experimentally show our method can provide considerable improvements within 20 to 50 refinement steps.

\begin{table*}[ht!]
\small
\centering
\begin{tabular}{x{2.8cm}|x{1.9cm}|x{1.9cm}|x{1.9cm}|x{1.9cm}|x{1.9cm}|x{1.9cm}}
\toprule
{} &  \multicolumn{2}{c}{$\Delta$ Good} & \multicolumn{2}{c}{$\Delta$ KL} & \multicolumn{2}{c}{$\Delta$ JS} \\
Metric & Pearson & Spearman & Pearson & Spearman & Pearson & Spearman \\
\midrule
Z \cite{dawid1997prequential} &  0.02 (0.72) & 0.11 (0.52) & 0.15 (0.30) & -0.01 (0.37) & 0.07 (0.41) & -0.06 (0.35) \\
ECE (Naeini 2015) & -0.18 (0.54) & -0.16 (0.55) & 0.32 (0.25) & 0.20 (0.42) & 0.31 (0.35) & 0.24 (0.47) \\
BS \cite{brier1950verification} & \textbf{-0.68 (0.01)} & \textbf{-0.72 (0.01)} & \textbf{0.51 (0.04)} & 0.46 (0.12) & \textbf{0.77 (0.00)} & \textbf{0.77 (0.00)} \\
\bottomrule
\end{tabular}
\caption{Statistical correlation between the performance gain from the rejection step and the score of different diagnostic metrics. We compute the Pearson's and Spearman's correlation coefficients based on the results of the MH method using discriminators fine-tuned for different iterations. Among the three diagnostic metrics, only the Brier Score has significant correlation with the performance gain ($p \le 0.05$).}
\label{tab:correlation}
\end{table*}

\begin{figure*}[t]
    \centering
    \includegraphics[scale=\szcelebbox]{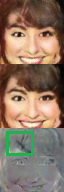}\includegraphics[scale=\szcelebbox]{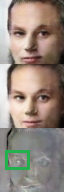}\includegraphics[scale=\szcelebbox]{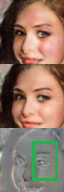}\includegraphics[scale=\szcelebbox]{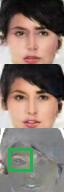}\includegraphics[scale=\szcelebbox]{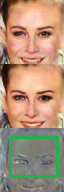}\includegraphics[scale=\szcelebbox]{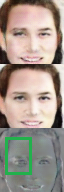}\includegraphics[scale=\szcelebbox]{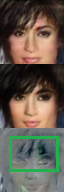}\includegraphics[scale=\szcelebbox]{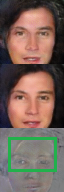}\includegraphics[scale=\szcelebbox]{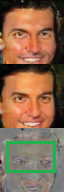}
    \caption{Qualitative results of our collaborative sampling method for DCGAN on the CelebA at $64 \times 64$ resolution. The DCGAN model is first trained for 30 epochs. The discriminator is further shaped for one epoch. The generated samples are refined at the 2nd layer of the generator for 50 steps. \textit{(Top)} Samples produced by the generator. \textit{(Middle)} Samples produced by our collaboratively sampling method. \textit{(Bottom)} The differences between the generated and refined images are highlighted for visualization.}
    \label{fig:image}
\end{figure*}

\section{Experiments}

In this section, we present experimental results to validate the proposed collaborative sampling scheme. We first show that our method outperforms the existing sampling methods on several GAN variants for modeling data distributions. Moreover, we demonstrate the benefits of our method in an image manipulation task, for which the previous accept-reject samplers are not applicable. Finally, we examine the effect of discriminator shaping as well as the choice of the refinement layer. 

\subsection{Synthetic Data}

We first evaluate our collaborative sampling scheme on a synthetic 2D dataset, which comprises of an imbalanced mixture of 8 Gaussians. 90\% of the real samples are drawn from two Gaussian components while the rest 10\% are drawn from the other six. We use a standard fully-connected MLP with 6 hidden layers and 64 hidden units per layer to model the generator and the discriminator. We shape the discriminator for $5k$ additional iterations after terminating the standard GAN training and conduct a maximum $50$ sample refinement steps in the data space with a step size of $0.1$. For fair comparison, we set the hyperparameter $\gamma$ in the rejection sampling method \cite{azadi2018discriminator} to 1.0 and the MC iteration number $k$ in the Metropolis-Hasting method \cite{turner2019metropolis} to 20. In addition, for fairness with regards to our discriminator shaping, we train the discriminator for $5k$ additional iterations before running these accept-reject methods.

Figure~\ref{fig:2D} shows the qualitative results of different sampling methods. The standard GAN training gradually runs into mode collapse on the imbalanced dataset, resulting in high sample quality but low diversity after $9k$ iterations. On the other hand, if the training procedure is early stopped ($1k$ iterations), the obtained generator can neither produce realistic samples nor provide complete coverage of the real data. The previous accept-reject sampling methods applied to the generator at this stage can successfully reject the majority of the bad samples, but fail to recover the real distribution. In contrast, our collaborative sampling scheme succeeds in obtaining samples of both high quality and high diversity.

We next evaluate our method quantitatively, following the previous protocol in \cite{azadi2018discriminator,turner2019metropolis}. Samples that are less than four standard deviations away from the nearest Gaussian component are considered as \textit{good}. We compute the KL divergence between the categorical distributions of real samples and good generated samples to measure the diversity of the good generated samples. To evaluate the overall performance, we introduce an extra category for the bad samples and compute the JS divergence between the augmented categorical distributions of real samples and all generated samples. As shown in Figure~\ref{fig:2D_eval}, our method exhibits superior performance in comparison to the existing sampling methods. 

Figure~\ref{fig:diagnosis} shows the scores of the diagnostic metrics as well as the JS divergence resulting from the MH algorithm using discriminators  fine-tuned for different iterations. In addition to the Z-statistic promoted in \cite{turner2019metropolis}, we also take the expected calibration error (ECE) \cite{naeini2015obtaining}, another popular metric for neural network calibration \cite{guo2017calibration} into comparison. It is visually apparent that the evolvement of the Z-statistic is dramatically different from the other metrics. More detailed correlation coefficients between the diagnostic metrics and then performance gains from the rejection step are summarized in Table~\ref{tab:correlation}. Compared with the Z-statistic and ECE, the Brier Score exhibits a significantly stronger correlation with the performance gains, which validates the effectiveness of the Brier Score for discriminator diagnosis. 

\begin{figure}[ht]
\centering
    \begin{subfigure}[b]{0.495\columnwidth}
        \includegraphics[width=0.95\columnwidth]{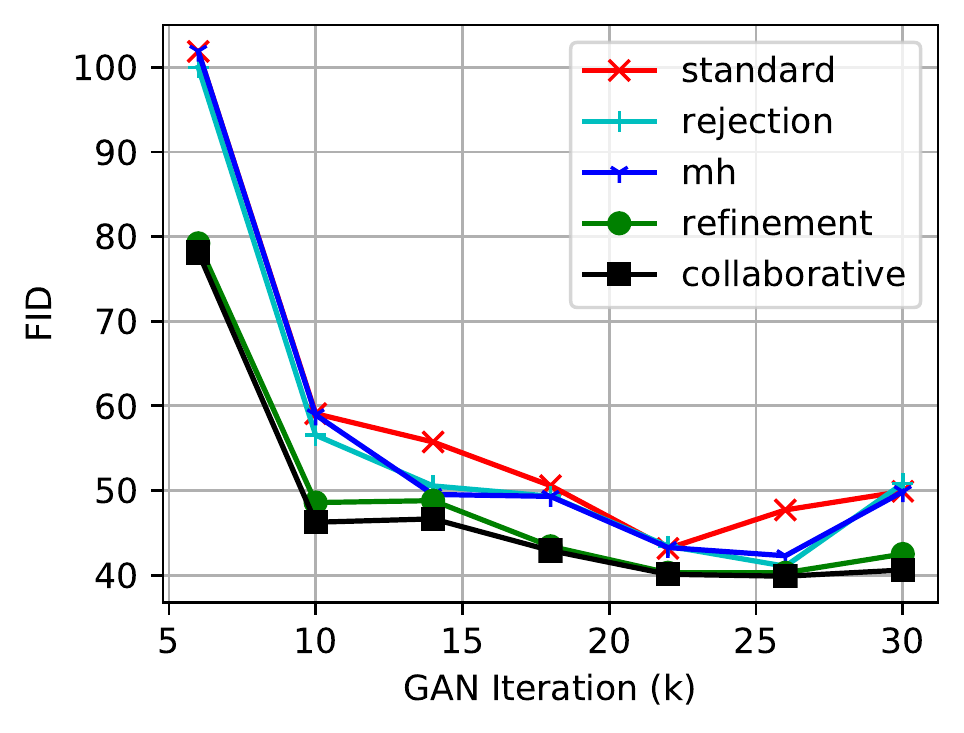}
        \caption{CIFAR10}
        \label{fig:cifar_is}
    \end{subfigure}
    \begin{subfigure}[b]{0.495\columnwidth}
        \includegraphics[width=0.95\columnwidth]{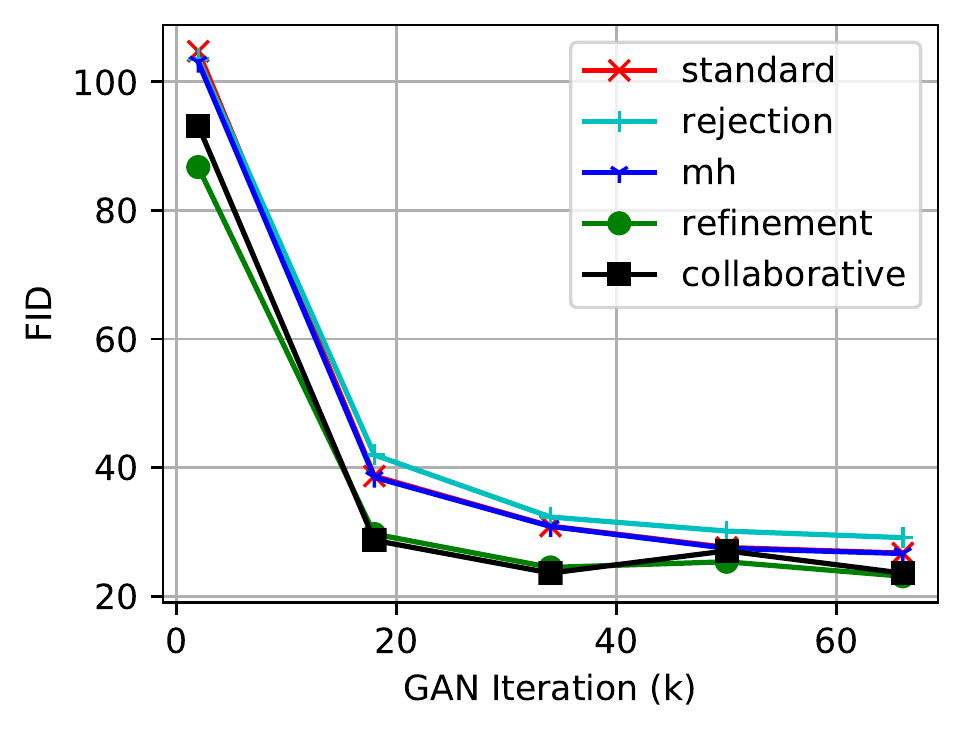}
        \caption{CelebA}
        \label{fig:celeb_is}
    \end{subfigure}
\caption{Quantitative comparison between our collaborative sampling scheme and baseline sampling methods for DCGAN on CIFAR10 and CelebA. Lower is better for FID.}
\label{fig:score_cifar_celeba}
\end{figure}

\begin{figure}[ht]
\centering
    \begin{subfigure}[b]{0.495\columnwidth}
        \includegraphics[height=0.14\textheight]{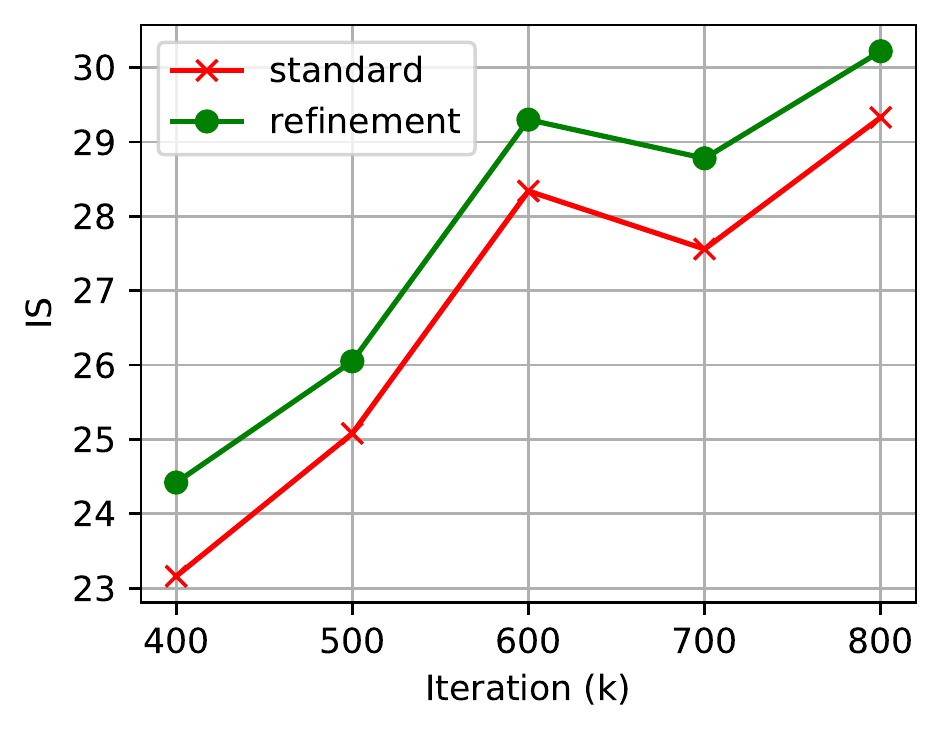}
        \caption{IS}
        \label{fig:sgan_cifar_is}
    \end{subfigure}
    \begin{subfigure}[b]{0.495\columnwidth}
        \includegraphics[height=0.14\textheight]{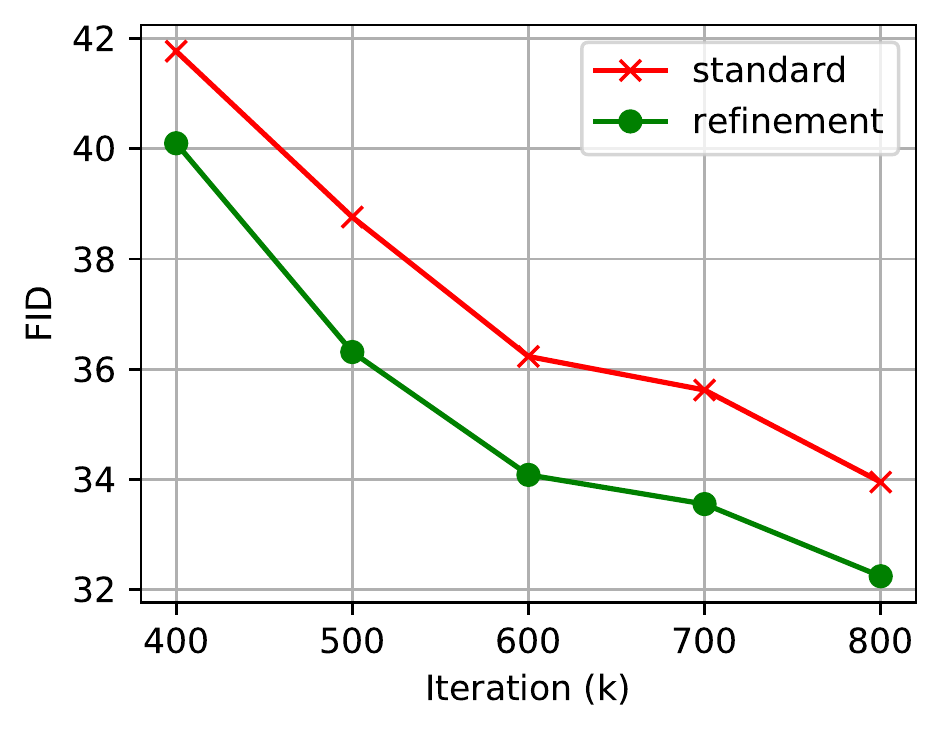}
        \caption{FID}
        \label{fig:sgan_celeb_is}
    \end{subfigure}
    \caption{Quantitative comparison between our collaborative sampling scheme and the standard sampling for SAGAN on ImageNet (Batch size 16). Higher is better for IS and lower is better for FID.}
    \label{fig:score_imagenet}
\end{figure}

\subsection{Image Generation} \label{sec:exp_image}

We next demonstrate the efficacy of our method in image generation tasks. In our experiments, we use the standard DCGAN \cite{radford_unsupervised_2015} for modelling the CIFAR10 \cite{krizhevsky2009learning} and the CelebA \cite{liu2015deep} datasets, and the SAGAN \cite{zhang2019self} for modelling ImageNet \cite{deng_imagenet:_2009} at $128\times128$ resolution. For sample refinement, we conduct a maximum of 50 refinement steps with a step size of 0.1 in a middle layer of the generator for the DCGAN and 16 updates with a step size of 0.5 for the SAGAN. Performance is quantitatively evaluated using the Inception Score (IS) \cite{salimans_improved_2016} and the Fr\'echet Inception Distance (FID) \cite{heusel2017gans} on 50k images.

As shown in Figure~\ref{fig:score_cifar_celeba} and Figure~\ref{fig:score_imagenet}, our collaborative sampling scheme provides consistent performance boost at each training stage across different datasets and GAN variants, suggesting the strong ability of our method to improve the model distribution of complex data. In addition to the quantitative improvements, Figure~\ref{fig:image} qualitatively compares the images proposed by the generator and those produced by our method on the CelebA dataset. The perceptual differences between the generated and refined images demonstrate the effectiveness of our method in identifying artifacts and improving image quality. 

\begin{figure}[ht!]
    \centering
    \includegraphics[scale=\szcyclebox]{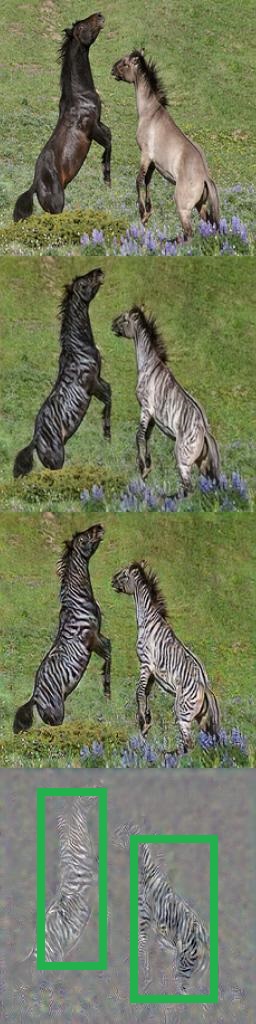}\includegraphics[scale=\szcyclebox]{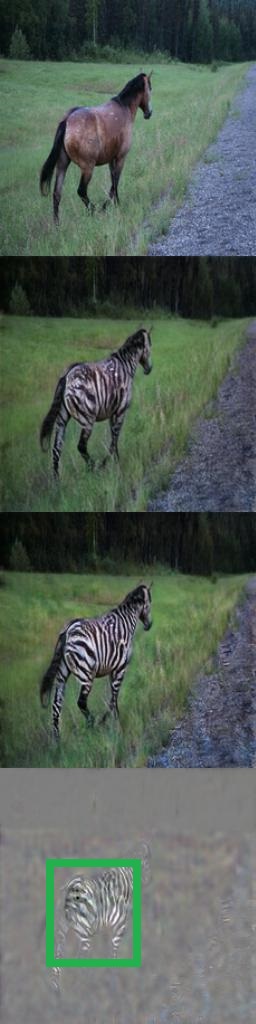}\includegraphics[scale=\szcyclebox]{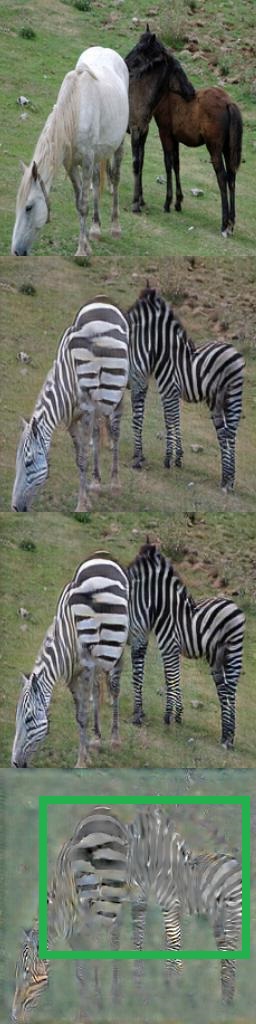}\includegraphics[scale=\szcyclebox]{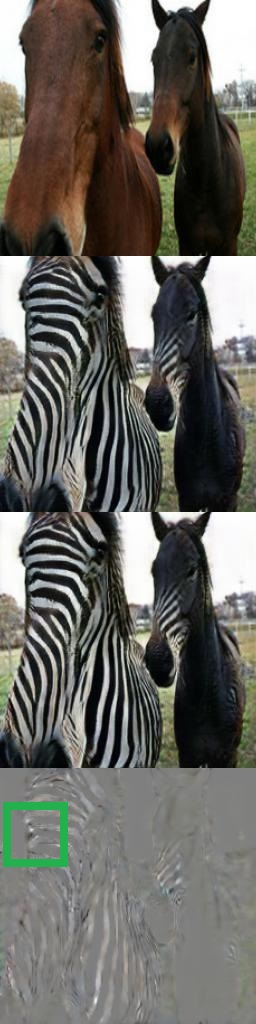}
    \caption{Results of our collaborative sampling scheme in CycleGAN for unpaired image-to-image translation at $256 \times 256$ resolution. The real horse images (top) are translated into synthetic zebra images (second row), which are further refined by our method (third row). The differences between the translated and refined images are concentrated on zebra patterns (bottom).} 
    \label{fig:cyclegan}
\end{figure}

\subsection{Image Manipulation}

We next evaluate our collaborative sampling scheme with CycleGAN \cite{zhu2017unpaired}, a popular method for image-to-image translation. To improve the image quality, we perform a maximum of 100 refinement steps. As shown in Figure~\ref{fig:cyclegan}, the modifications made by our method at $256\times256$ resolution concentrates on the pattern of the target class without affecting background semantics. This result validates the unique advantage of our method over the rejection algorithms for enhancing the output quality in the image manipulation tasks. 

\subsection{Key Attributes}

We finally investigate the impact of two key attributes of our method through experiments on the MNIST \cite{lecun1998gradient}. Here, we use the original NS-GAN \cite{Goodfellow2014GenerativeAN} as a baseline and apply our collaborative sampling scheme for $20$ refinement steps with a step size of $0.1$.

\begin{figure}[t]
    \centering
        \includegraphics[scale=\szablation]{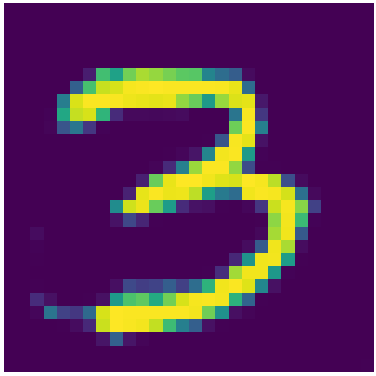}
        \includegraphics[scale=\szablation]{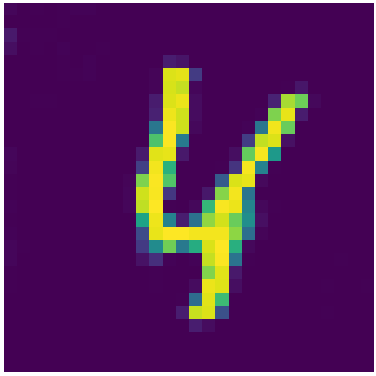}
        \includegraphics[scale=\szablation]{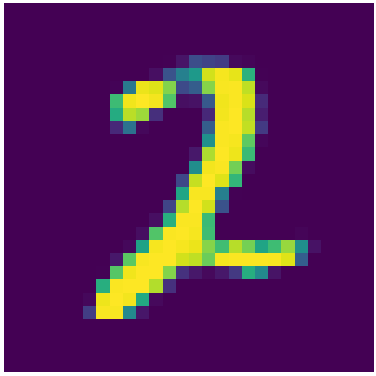}
        \includegraphics[scale=\szablation]{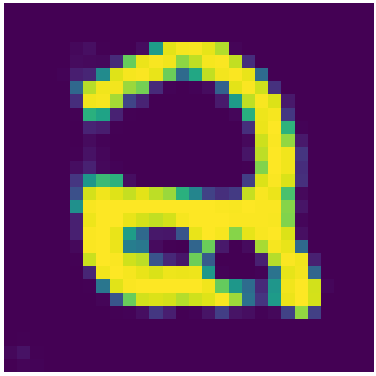}
        \includegraphics[scale=\szablation]{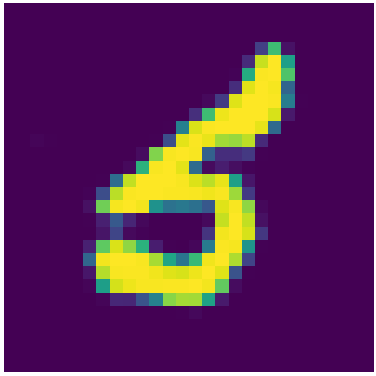}
        \\
        \includegraphics[scale=\szablation]{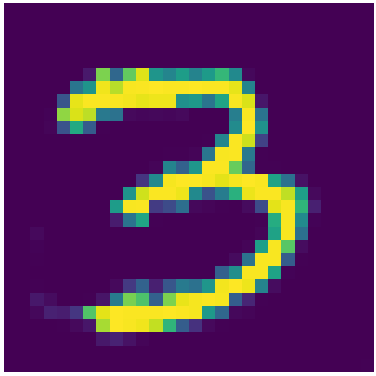}
        \includegraphics[scale=\szablation]{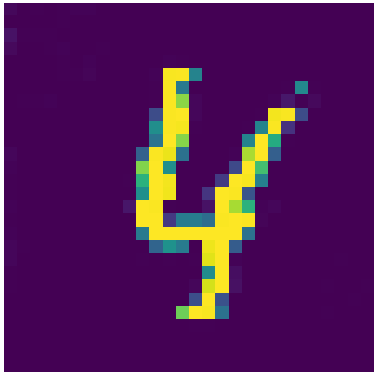}
        \includegraphics[scale=\szablation]{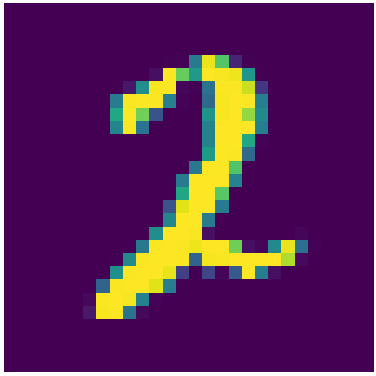}
        \includegraphics[scale=\szablation]{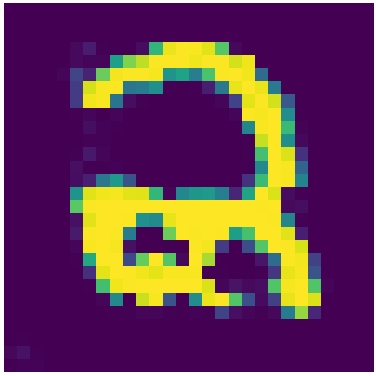}
        \includegraphics[scale=\szablation]{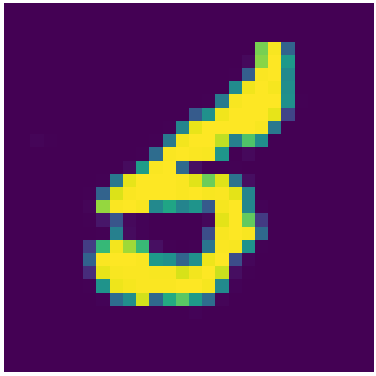}
        \\
        \includegraphics[scale=\szablation]{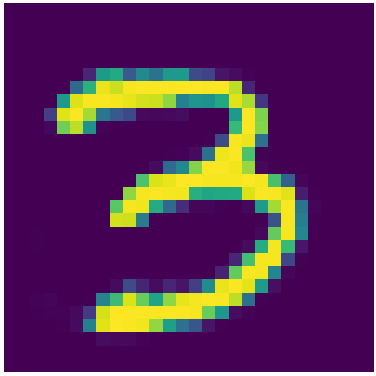}
        \includegraphics[scale=\szablation]{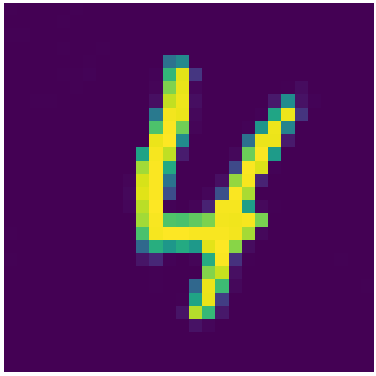}
        \includegraphics[scale=\szablation]{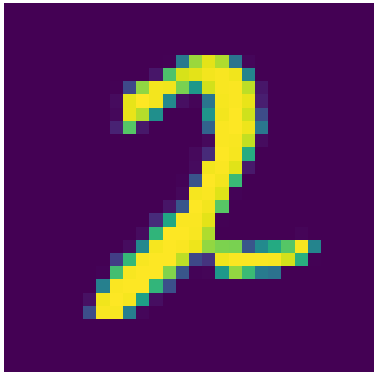}
        \includegraphics[scale=\szablation]{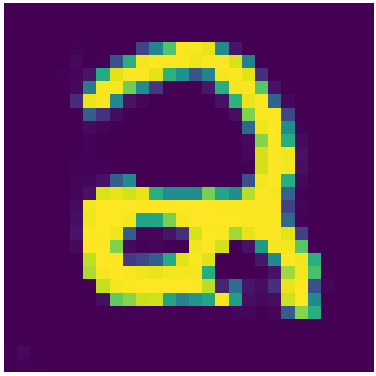}
        \includegraphics[scale=\szablation]{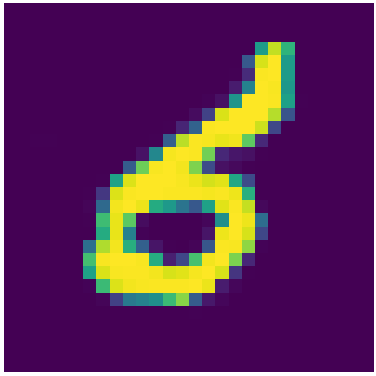}
    \caption{Qualitative effect of discriminator shaping. Refining images proposed by the generator \textit{(first row)} using the standard discriminator without additional shaping leads to worse images \textit{(second row)} in comparison to the images obtained after discriminator shaping \textit{(third row)}.} 
    \label{fig:d_shaping}
\end{figure}
\begin{figure}[t]
    \centering
    \begin{subfigure}[b]{0.495\columnwidth}
        \includegraphics[width=\columnwidth]{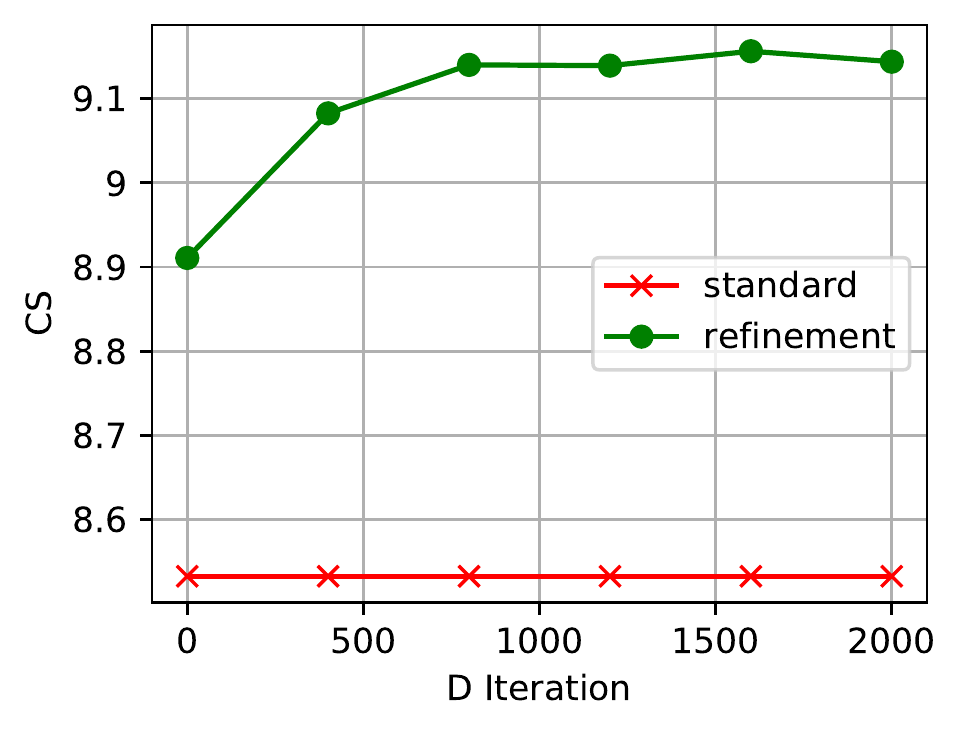}
        \caption{CS}
        \label{fig:shape_is}
    \end{subfigure}
    \begin{subfigure}[b]{0.495\columnwidth}
        \includegraphics[width=\columnwidth]{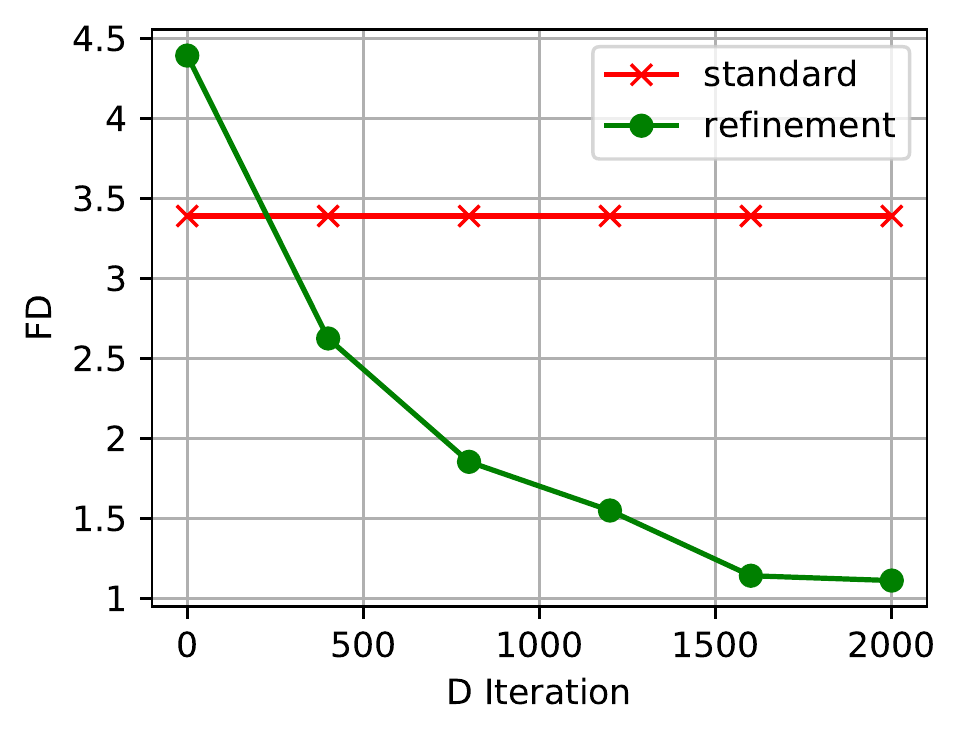}
        \caption{FD}
        \label{fig:shape_fd}
    \end{subfigure}
\caption{Quantitative effect of discriminator shaping. We apply our sample refinement method using discriminators shaped for different iterations. Higher is better for CS and lower is better for FD.}
\label{fig:curve_shaping}
\end{figure}

\begin{figure}[t]
    \centering
        \includegraphics[scale=\szablation]{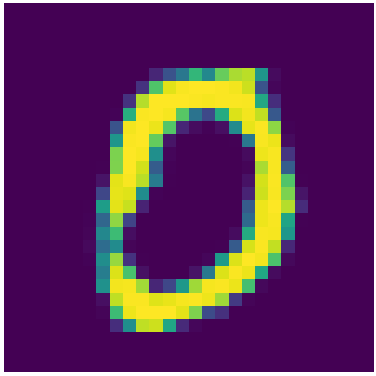}
        \includegraphics[scale=\szablation]{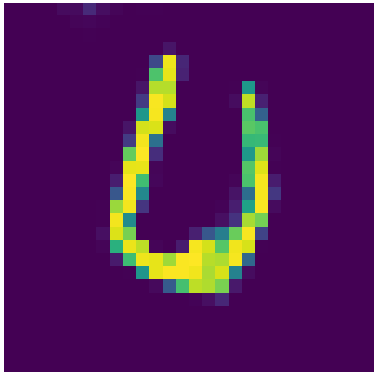}
        \includegraphics[scale=\szablation]{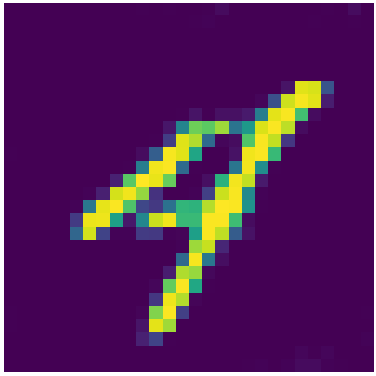}
        \includegraphics[scale=\szablation]{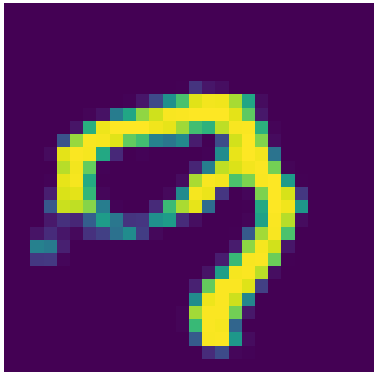}
        \includegraphics[scale=\szablation]{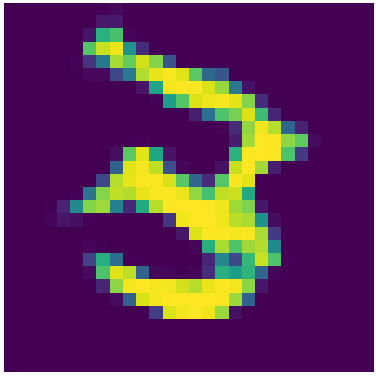}\\
        \includegraphics[scale=\szablation]{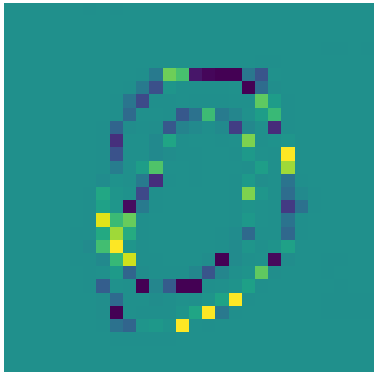}
        \includegraphics[scale=\szablation]{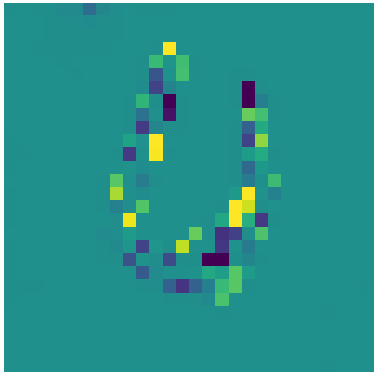}
        \includegraphics[scale=\szablation]{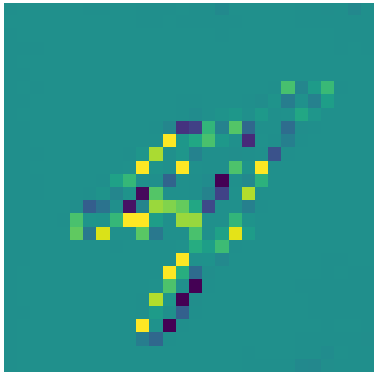}
        \includegraphics[scale=\szablation]{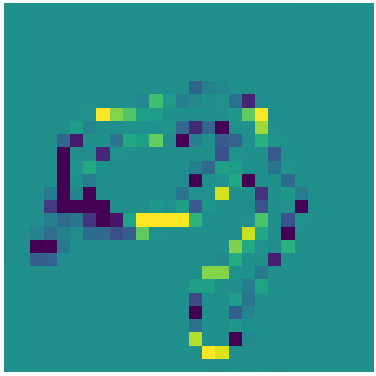}
        \includegraphics[scale=\szablation]{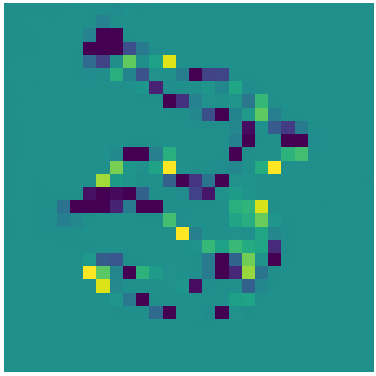}\\
        \includegraphics[scale=\szablation]{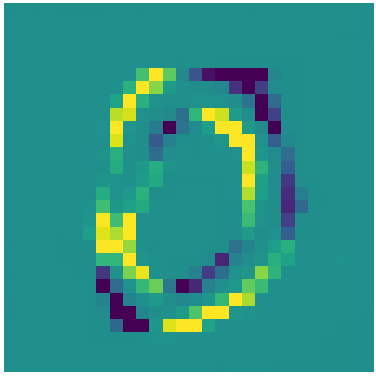}
        \includegraphics[scale=\szablation]{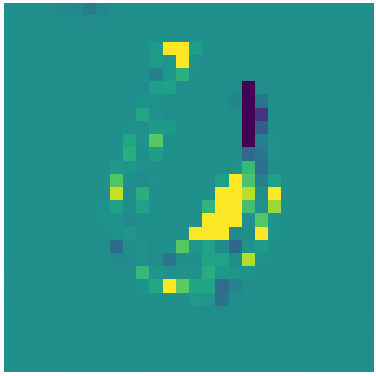}
        \includegraphics[scale=\szablation]{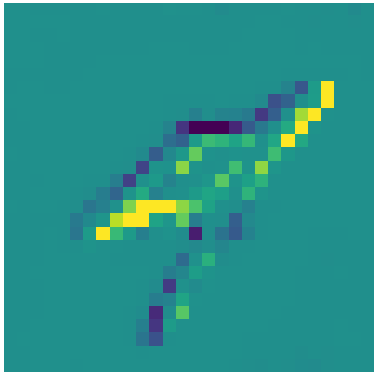}
        \includegraphics[scale=\szablation]{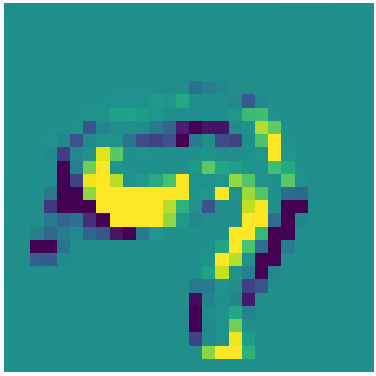}
        \includegraphics[scale=\szablation]{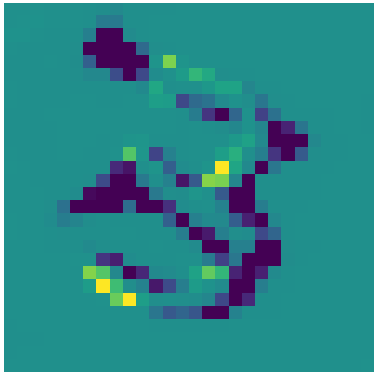}\\
        \includegraphics[scale=\szablation]{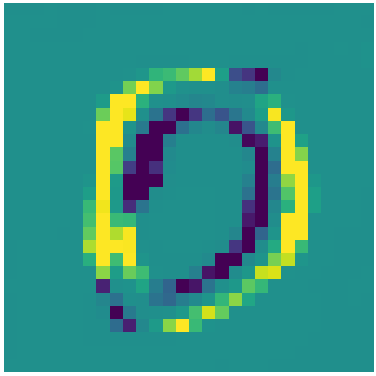}
        \includegraphics[scale=\szablation]{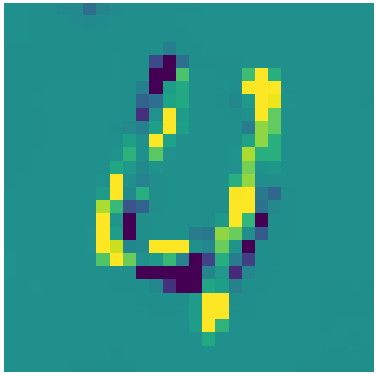}
        \includegraphics[scale=\szablation]{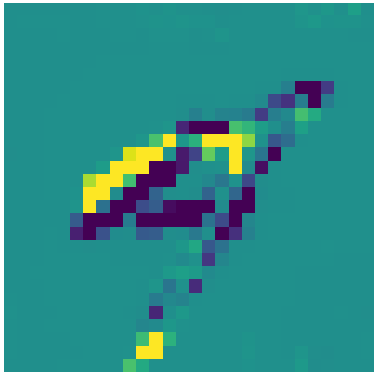}
        \includegraphics[scale=\szablation]{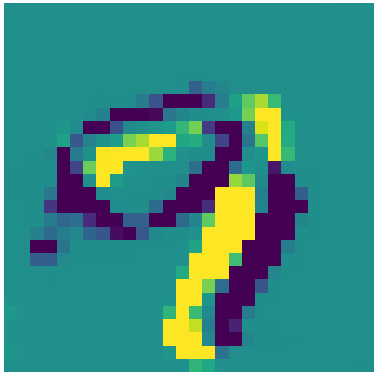}
        \includegraphics[scale=\szablation]{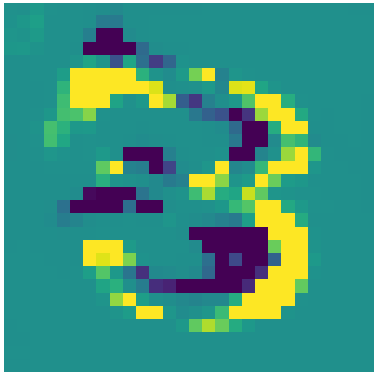}
    \caption{Qualitative effect of the choice of refinement layer. The last three rows show the differences between the generated sample (first row) and the refined samples when sample refinement (i) performed at the output layer (second row), (ii) the middle layer ( third row), and (iii) the input of the generator (last row). We can observe sample modifications from micro to macro scales.}
    \label{fig:layerwise}
\end{figure}

\subsubsection{Effect of Discriminator Shaping}

We highlight the importance of the proposed discriminator shaping by qualitatively comparing the MNIST images produced by three different sampling schemes: (a) standard GAN sampling (b) collaboratively sampling without discriminator shaping and (c) collaborative sampling with discriminator shaping. As shown in Figure~\ref{fig:d_shaping}, when the generated images contain small artifacts, the refinement process guided by the standard discriminator fails to remove the artifacts and instead adds more noises.
In contrast, the shaped discriminator leads to visually more realistic digits. Figure~\ref{fig:curve_shaping} shows the quantitative effect of discriminator shaping on classifier score (CS) and Fr\'echet distance (FD), reaffirming the necessity of discriminator shaping. 

\subsubsection{Effect of Refinement Layer}

To examine the impact of the choice of the refinement layer, we visualize the difference between the refined samples and the originally proposed samples as a function of the layer index in Figure~\ref{fig:layerwise}. The sample refinement performed at the output layer results in local modifications, whereas the refinement at the low-level activation map alters the global semantics. The choice of the middle layer leads to a balanced performance, fixing the local artifacts in ``0" and ``3" while making global changes in the other images that are far from being realistic. 


\section{Related Work}

Designing collaborative mechanisms in addition to adversarial training has garnered growing interest in the past couple of years. \cite{Yann2016GCN} promoted to replace the discriminator by a collaborator to provide encouraging feedback. Recent works \cite{xie2018cooperative,chen2019self,seddik_generative_2019} proposed concrete methods for training generative models collaboratively. In contrast, our work is focused on the design of collaboration mechanism during the sampling process.

To obtain desired samples, one line of work \cite{zhu2016generative,nguyen2017plug,Yeh2016SemanticII,samangouei_defense-gan:_2018} employed gradient-based optimization in the latent space of GANs. The goal of these methods is essentially to seek samples from the learned data manifold closest to the target, whereas we aim to improve the learned data prior. By performing sample refinement at a selected layer of the generator, our method enables the refined samples to go beyond the data manifold modelled by the generator. 

Another line of recent work proposed to modify the activation in a middle layer of the generator in order to manipulate samples with human intervention \cite{bau2019gandissect} or an additional neural network \cite{shama2019adversarial}. On the contrary, our work provides a generic sample refinement method guided by the gradient from the discriminator, allowing one to exploit the full knowledge of the learned networks with theoretical inspirations. 

Our proposed discriminator shaping method can be viewed as a form of adversarial training, which is typically used to improve the robustness of a classifier \cite{madry_towards_2017}. Recently, \cite{zhou2018dont} proposed to train the discriminator adversarially in a restricted region for stabilizing the GAN training. Another concurrent work \cite{Santurkar2019ComputerVW} demonstrated the generative power of a single adversarially robust classifier through extensive experiments. Our method adopts the adversarial training approach with the goal of guiding the collaborative sampling process effectively at test time. 

\section{Conclusions}

We present a novel collaborative sampling scheme for using GANs at test time. Rather than disregarding the discriminator, we propose to continue using the gradients provided by a shaped discriminator to refine the generated samples. This is advantageous when the model distribution does not match the real data distribution. It is also highly valuable for applications where sample quality matters or the rejection sampling approach is not admissible. Orthogonal to existing techniques in GAN training, our method offers an additional degree of freedom to improve the generated samples empowered by the discriminator. 

\section{Acknowledgements}
We thank Sven Kreiss and Tao Lin for helpful discussions. We also thank Taylor Mordan, Dhruti Shah for valuable feedback on drafts of this paper. 

{\small
\bibliographystyle{aaai}
\bibliography{vita-G2N,yuejiang}
}

\end{document}